\documentclass[11pt, letterpaper]{article}  

\usepackage[left=1in, right=1in, top=1in, bottom=1in]{geometry}
\usepackage[utf8]{inputenc}
\usepackage[T1]{fontenc}
\usepackage{bm}
\usepackage{type1cm}
\usepackage{lettrine}
\usepackage{amsmath,amssymb,amsthm}
\usepackage{moreverb}
\usepackage{mathtools}
\usepackage{amsmath}
\usepackage{amssymb}
\usepackage{algorithmic}
\usepackage{graphics}
\usepackage{graphicx}
\usepackage{placeins}
\usepackage{subfigure}
\usepackage[figurename=Figure,font=footnotesize,labelfont=bf]{caption}
\usepackage{extarrows}
\usepackage[dvipsnames,table]{xcolor}
\usepackage{framed}
\usepackage{wrapfig}
\usepackage{bm}
\usepackage{mathrsfs}
\usepackage{mathabx}
\usepackage{multirow}
\usepackage{longtable}
\usepackage[bookmarks=true]{hyperref}
\usepackage{paralist}
\usepackage{indentfirst}
\usepackage{relsize}
\usepackage{extarrows}
\usepackage{upgreek}
\usepackage{bm}
\usepackage{mwe}
\usepackage{booktabs}
\usepackage{authblk}
\usepackage{lettrine}
\usepackage{type1cm}
\usepackage{threeparttable}
\usepackage[sort&compress,numbers]{natbib}
\usepackage[normalem]{ulem}

\graphicspath{ {./Figure/} }
\newcommand{\eref}[1]{(\ref{#1})}
\providecommand{\keywords}[1]{\textbf{\textit{Keywords: }} #1}

\hypersetup{
bookmarksopen=true,
bookmarksnumbered=true,
unicode=false,
pdftoolbar=true,
pdfmenubar=true,
pdffitwindow=false,
pdfstartview={FitH},
pdftitle={Nowcast3D for effective precipitation nowcasting},
pdfauthor={Huaguan Chen, Wei Han, Haofei Sun, Ning Lin, Xingtao Song, Yunfan Yang, Jie Tian, Yang Liu, Ji-Rong Wen, Xiaoye Zhang, Xueshun Shen, Hao Sun},
pdfkeywords={Precipitation nowcasting, 3D volumetric radar reflectivity, physics-guided learning},
pdfnewwindow=true,
colorlinks=true,
linkcolor=black,
citecolor=black,
filecolor=black,
urlcolor=black
}

\ifdefined\pdfminorversion\pdfminorversion=7\fi
\begin{document}

\title{{\textbf{Nowcast3D for effective precipitation nowcasting}}}

\author[1]{Huaguan Chen}
\author[2,3$^*$]{Wei Han}
\author[4]{Haofei Sun}
\author[1]{Ning Lin}
\author[5]{Xingtao Song}
\author[6,7]{Yunfan Yang}
\author[2,8]{\\Jie Tian}
\author[9]{Yang Liu}
\author[1,10]{Ji-Rong Wen}
\author[3,11]{Xiaoye Zhang}
\author[2,3]{Xueshun Shen}
\author[1,10$^*$]{Hao Sun\vspace{6pt}}

\affil[1]{\small Gaoling School of Artificial Intelligence, Renmin University of China, Beijing, China}
\affil[2]{\small Earth System Modeling and Prediction Center, China Meteorological Administration, Beijing, China} 
\affil[3]{\small State Key Laboratory of Severe Weather Meteorological Science and Technology, Beijing, China}
\affil[4]{\small Shanghai Typhoon Institute, China Meteorological Administration, Shanghai, China}
\affil[5]{\small School of Atmospheric Physics, Nanjing University of Information Science and Technology, Nanjing, China}
\affil[6]{\small Institute of Atmospheric Physics, Chinese Academy of Sciences, Beijing, China}
\affil[7]{\small College of Earth and Planetary Sciences, University of Chinese Academy of Sciences, Beijing, China}
\affil[8]{\small Chinese Academy of Meteorological Sciences, China Meteorological Administration, Beijing, China}
\affil[9]{\small School of Engineering Science, University of Chinese Academy of Sciences, Beijing, China}
\affil[10]{\small Beijing Key Laboratory of Research on Large Models and Intelligent Governance, Beijing, China}
\affil[11]{\small Laboratory of Climate Change Mitigation and Carbon Neutrality, Henan University, Zhengzhou, China\vspace{18pt}}
\affil[*]{Corresponding authors}

\date{}

\maketitle

\normalsize

\vspace{-12pt} 

\begin{abstract}
	\small 
{
The evolution of extreme precipitation depends on height-varying wind fields and coupled physical processes, underscoring the importance of three-dimensional information for precipitation nowcasting. Radar volume scans provide such an observational basis. Most learning-based systems, however, either compress these scans into two-dimensional products, losing vertical information needed for height-dependent storm evolution, or rely on purely data-driven mappings that inadequately represent the physical evolution of extreme precipitation. We introduce Nowcast3D, a gray-box learning framework for 3D radar-volume nowcasting, which integrates physically structured neural operators with a conditional diffusion model. The neural operators produce a physics-guided forecast via 3D advection, local diffusion and microphysics processes. Conditioned on this forecast, the diffusion model generates uncertainty-quantified ensembles. Trained on provincial-scale radar volumes covering a $10.24^\circ \times 10.24^\circ$ region and fine-tuned on a $2.56^\circ \times 2.56^\circ$ city-scale region ($0.01^\circ \approx 1$ km), Nowcast3D produces effective near-real-time reflectivity forecasts up to 3 hours ahead. It consistently outperforms competitive baselines in both cross-region and temporal-out-of-sample evaluations, and its inferred motion fields show good consistency with collocated wind-profiler observations. In a blind evaluation with 160 meteorologists, Nowcast3D ranked first and was preferred in 57\% of post-hoc assessments, compared with 27\% for the leading baseline. These results highlight its operational potential for effective nowcasting of extreme precipitation events.}
\end{abstract}


\vspace{12pt} 

\section*{Introduction}
On 16 June 2024, extreme rainfall in Meizhou, Guangdong Province, China triggered severe floods and landslides, affecting more than 160,000 people and leaving 55 individuals dead or missing. Such events starkly illustrate the severe threat posed by intense precipitation to lives and infrastructure. They also highlight the pressing need for nowcasting systems capable of delivering higher accuracy, in order to facilitate timely warnings and improve risk mitigation.

Accurate short-term precipitation forecasting is critical for pre-disaster planning. Weather radar provides an observational foundation well-suited to this task. Modern radar networks acquire three-dimensional (3D) reflectivity fields with a horizontal resolution approaching 0.01° ($\approx$ 1 km), minute-scale temporal sampling, and vertical spacing of a few hundred meters, yielding dense volumetric measurements of precipitating systems \cite{starzec2018using,franch2020taasrad19,otsuka2016precipitation}. Radar reflectivity is closely related to instantaneous precipitation intensity. Consequently, time sequences of 3D radar data capture not only the movement of precipitation patterns but also the vertical structure of reflectivity. Variations in echo-top height, bright-band signatures, and layer-dependent reflectivity are closely linked to the initiation, growth, and propagation of convection, which strongly governs the evolution and organization of heavy precipitation \cite{steiner1995climatological,biggerstaff2000improved}.

Numerical weather prediction (NWP) remains indispensable from synoptic to subseasonal scales, as it explicitly integrates the governing equations of atmospheric motion and thermodynamics. {Modern rapid-refresh NWP systems provide frequent updates. For minute-scale convective nowcasting, their practical use is still affected by the cadence of convective-scale data assimilation, forecast-production and dissemination latency, and uncertainty in kilometer-scale initial conditions \cite{sun2014use}. These factors affect the direct use of NWP guidance for minute-scale convective nowcasting at radar-observed scales.}
{Recent data-driven weather forecasting models have achieved substantial gains in prediction speed and scalability \cite{bi2023accurate,lam2023learning,kochkov2024neural,bodnar2025foundation,sun2025data,cui2025forecasting,allen2025end}. Much of this progress has focused on global or broad-scale atmospheric prediction. Radar-based convective nowcasting poses a different problem. It requires kilometer-scale, minute-frequency prediction of three-dimensional echo organization, growth and decay. The target is the rapidly changing structure and intensity of radar-observed convective echoes, rather than the evolution of large-scale atmospheric state variables alone.}

Radar-observation-based forecasting methods remain the dominant paradigm for precipitation nowcasting. {Optical-flow kinematic extrapolation methods}, such as pySTEPS \cite{pulkkinen2019pysteps}, infer motion from recent fields and then advect them forward in time. {These methods do not explicitly model the thermodynamic and microphysical processes driving intensity changes.}
Although these methods can preserve {large-scale transport information} and achieve good skill at very short lead times, their positional and intensity errors grow rapidly as the lead time increases. End-to-end data-driven models trained on extensive radar archives {have shown great potential}, e.g.,
 DGMR \cite{ravuri2021skilful}, SimVP \cite{gao2022simvp}, among others \cite{shi2015convolutional,wang2017predrnn,gao2022earthformer,espeholt2022deep}. However, in the absence of explicit physical constraints, their forecasts tend to over-smooth and weaken intense convective features, limiting their reliability for applications that depend on accurate prediction of extreme events.

Hybrid learning approaches that embed physical processes into neural networks have emerged to narrow this gap \cite{zhang2023skilful,das2024hybrid}. For instance, NowcastNet \cite{zhang2023skilful} unifies physical-evolution schemes and conditional-learning methods, thereby anchoring forecasts to transport dynamics and enhancing the physical coherence of prediction. {The recent medium-range weather model PARADIS \cite{paradis2026} further illustrates the utility of physically structured learning systems for weather prediction.} However, notable limitations remain. Many existing hybrid methods operate on 2D radar composites, most commonly the column-maximum reflectivity (MAX$(Z)$). This discards critical information on the vertical structure of atmosphere and therefore fails to capture features that are central to many convective and frontal systems \cite{franch2020taasrad19,starzec2018using}, such as height-dependent wind shear and differential motion between layers or echo-top evolution. 
{Although some purely data-driven approaches directly leverage 3D radar observations and retain height-resolved radar information \cite{kim2021improving,sun2022convective,baron2023nowcasting,wan2026generative}, they may still produce nowcasts with unrealistic echo motion and intensity evolution, as well as large location errors, particularly at longer lead times.}

To address these limitations, we develop Nowcast3D, a gray-box nowcasting network that unifies physics-inspired neural operators and a conditional diffusion model, operating directly on 3D volumetric radar reflectivity. By modeling convective systems in 3D space, Nowcast3D explicitly {resolves the height-dependent atmospheric structure} and its temporal evolution. Specifically, the model learns to decompose the observed reflectivity dynamics into three physically interpretable fields. {A 3D flow field governs advective transport, while a spatially varying anisotropic diffusion field captures local dispersive spreading. A residual source term accounts for unresolved microphysical and convective contributions to local echo growth and decay}. These components form a physics-grounded yet learnable deterministic core that advances forecasts under explicit constraints.
To correct for prediction errors and represent uncertainty, the deterministic forecast produced by this core, together with the input radar observations, conditions a diffusion model.
The diffusion model then generates ensemble predictions by sampling storm structure and localized intensity.
{Together, the deterministic forecast and generative refinement help reduce over-smoothing and the underestimation of extremes. Conditioning generation on the physics-informed forecast also promotes physical consistency.}

{Here, we use ``gray-box learning'' to denote the forecasting model, where physically interpretable process operators are embedded within the prediction pathway. Specifically, the radar reflectivity changes in Nowcast3D are represented by operators for advection, anisotropic local diffusion and residual source tendencies. This operator-level formulation exposes intermediate transport, spreading and intensity-change components that are otherwise hidden in ``black-box'' extrapolation models. This explicit process decomposition distinguishes Nowcast3D from most existing physics-informed learning approaches \cite{karniadakis2021physics}, in which physical information is often applied as a soft penalty that regularizes the loss function.}

Nowcast3D was trained on provincial-scale 3D radar volumes covering a $10.24^\circ \times 10.24^\circ$ region at $0.04^\circ$ resolution (approximately 3$\sim$4~km), and subsequently fine-tuned on data of a city-scale $2.56^\circ \times 2.56^\circ$ region at $0.01^\circ$ resolution (approximately 1~km), enabling near-real-time forecasting with lead times up to 3~hours. In cross-region generalization tests and temporal-out-of-sample evaluations, Nowcast3D demonstrates robust performance and consistently outperforms several competitive baselines. 
{Notably, the model can also infer motion fields without any labeled wind supervision. The comparison with wind-profiler radar observations confirmed the plausibility of the inferred transport dynamics.}
Finally, in a blind evaluation involving 160 meteorologists from 30 provinces (or municipalities) in China, the performance of Nowcast3D ranked first and was preferred in 57\% of post-hoc assessments, {highlighting its operational potential for extreme precipitation nowcasting.}

\begin{figure}[htbp]
\centering
\includegraphics[width=0.9\linewidth]{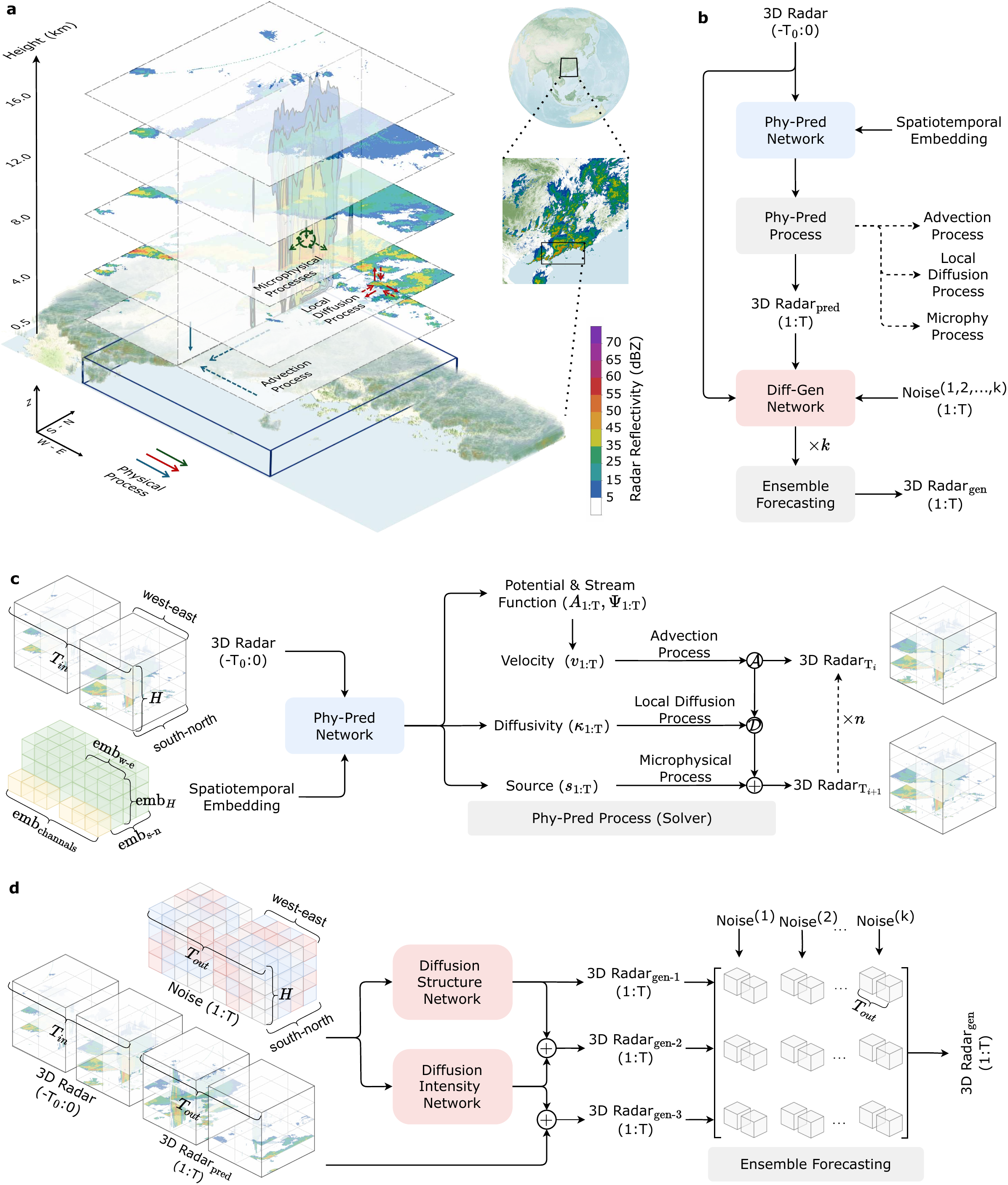}
\caption{\textbf{Schematic of the Nowcast3D model.}
\textbf{a}, Physical decomposition of reflectivity evolution. The changes in 3D radar reflectivity are decomposed into three contributions: advection by the 3D wind field, local diffusion capturing small-scale spreading, and microphysical tendencies that modify reflectivity in situ.
\textbf{b}, Overall model architecture. The Phy-Pred Network infers latent physical fields from input 3D radar with spatiotemporal embedding, and a physics-based predictor advances the system according to the three processes in \textbf{a}. The resulting physics forecast conditions a diffusion generator which produces an ensemble of probabilistic forecasts via drawing multiple independent noise realizations. Details of Nowcast3D are provided in Extended Data Figure~\ref{fig:all_model}.
\textbf{c}, Deterministic physics backbone (the Phy-Pred Process). 
The deterministic predictor performs each temporal update by explicitly applying advection, diffusion, and microphysical operators within the Phy-Pred Process module.
\textbf{d}, Diffusion-based probabilistic generation (the Diff-Gen Network). The diffusion model learns {physically structured} mesoscale patterns by the Diffusion Structure Network and stochastic residual components by the Diffusion Intensity Network. Under fixed conditioning, different sampled noise trajectories produce calibrated probabilistic 3D reflectivity forecasts.}
\label{fig:model}
\end{figure}

\section*{Results}

\subsection*{Nowcast3D model}

Nowcast3D is designed to forecast precipitation, in particular extreme events, based on volumetric radar reflectivity. The core idea is to decompose complex atmospheric evolution into physically interpretable components and establish {a unified learning architecture that sequentially combines a physics-inspired deterministic extrapolation network with a diffusion-based generative model} for probabilistic forecasts (see Figure~\ref{fig:model}). 

Our methodology decomposes the evolution of radar reflectivity into three dominant physical processes: advection, which captures the coherent drift of hydrometeors by 3D flow; local turbulent diffusion, representing the spreading of echoes toward neighboring grid points due to small-scale mixing; and a microphysical tendency that accounts for in-situ mass and phase changes (Figure~\ref{fig:model}\textbf{a}) \cite{sigrist2015stochastic}. {Here, diffusion should be interpreted as an effective radar-echo dispersion process caused by unresolved turbulent mixing and small-scale deformation, rather than molecular diffusion.} In this formulation, the advective field specifies, at each grid point, a dominant direction of motion, whereas the diffusive term relaxes the assumption of a single trajectory and allows reflectivity to fan out locally rather than move rigidly along a unique path. This advection–diffusion–microphysics decomposition provides a compact and interpretable description of reflectivity dynamics and serves as a strong prior for the deterministic core of Nowcast3D. 
{The overall architecture consists of two sequential stages} (Figure~\ref{fig:model}\textbf{b}): a physics-driven prediction module and a probabilistic generation module. The physics-driven module first produces a single, dynamically constrained deterministic forecast, which then serves as a reference state. The generative module constructs an ensemble of plausible future evolutions based on the reference, thereby capturing the intrinsic uncertainty of precipitation development. An overview of the full Nowcast3D pipeline is depicted in Extended Data Figure~\ref{fig:all_model}\textbf{a}, with detailed formulations presented in the \nameref{Methods} section.

The physics-driven module (Figure~\ref{fig:model}\textbf{c}) extrapolates the radar field using a neural network to infer the governing processes from a sequence of past observations and spatiotemporal embedding. In a single forward pass, an encoder processes the historical 3D radar volumes and outputs complete spatiotemporal fields of (i) an advective drift velocity, (ii) a local diffusion field, and (iii) a residual source term over the entire forecast horizon. Rather than explicitly reconstructing a sparse 3D velocity field, one branch of the network predicts scalar potential and vector stream-function fields, which are combined through a Helmholtz decomposition \cite{helmholtz1858integrale,cao2014application,xing2024HelmFluid} to produce a more spatially continuous drift velocity field. In parallel, other network components infer a spatially varying, anisotropic diffusivity tensor that governs the local spreading of echoes, and a source field representing the net effect of microphysical processes. As summarized in Eq.~\eref{eq:final-update}, the final deterministic forecast is obtained by numerically solving the advection–diffusion–source system defined by these three operators, using the last observed radar state as the initial condition and the network-inferred fields as the governing parameters. {Extended Data Figure~\ref{fig:all_model}\textbf{c} depicts the model overview. The complete Phy\mbox{-}Pred architecture is detailed in {Supplementary Figure~2} and {Supplementary Note~2.1}, while the spatiotemporal embedding configuration is provided in {Supplementary Note~1.4}.}

The probabilistic refinement stage is implemented via a conditional diffusion model with a dual-branch architecture (Figure~\ref{fig:model}\textbf{d}) \cite{ho2020denoising,song2020denoising,gong2024cascast,yu2024diffcast,price2025probabilistic}. {Diffusion models provide stable non-adversarial training and high-fidelity conditional sampling, enabling the preservation of sharp convective structures in probabilistic forecasts.} Both diffusion branches are conditioned on the same context, consisting of the past 3D radar volumes and the physics-driven deterministic forecast, so that the generative model remains anchored to the physically constrained evolution. The first branch (aka, the structure network) learns to generate a full 3D reflectivity field that captures the overall spatial and temporal organization of echoes over the forecast horizon. The second branch (aka, the intensity  network) learns a residual intensity field relative to the deterministic forecast, representing scale-localized amplification or weakening of precipitation intensity on top of the baseline solution. Using these two components, we can construct several types of samples: a purely generative structural realization, an intensity-refined deterministic forecast, and a sample that combines the generated structure with the learned intensity residual. As summarized in Eq.~\eref{eq:ensemble_construction}, drawing multiple independent noise vectors and propagating them through both diffusion branches under the shared conditioning yields a multi-member ensemble of 3D reflectivity forecasts. This provides a quantification of uncertainties in both spatial organization and intensity \cite{li2024generative}. {Extended Data Figure~\ref{fig:all_model}\textbf{d} shows an overview of Diff\mbox{-}Gen, with the complete architecture and implementation described in {Supplementary Figure~3} and {Supplementary Note~2.2}.}

{
The raw radar volumes firstly undergo mask construction, mask-guided low-level imputation and normalization (see Extended Data Figure~\ref{fig:all_model}\textbf{b}) before input into the model. Extended Data Figure~\ref{fig:radarcover} shows the radar coverage, the dominant near-surface missing-data patterns and representative examples of low-level completion. The imputation is applied only to missing reflectivity at lower elevation levels while keeping all observed echoes unchanged. This procedure provides spatially continuous radar inputs while avoiding modification of the measured signals. Details of the radar datasets, training-sample construction and preprocessing procedures are provided in \nameref{Methods} and {Supplementary Notes~1.2--1.3}.
}

{
For forecast-skill evaluation, Nowcast3D is compared with three representative baselines: pySTEPS \cite{pulkkinen2019pysteps}, SimVP \cite{gao2022simvp} and NowcastNet \cite{zhang2023skilful}. These methods represent extrapolation-based, purely data-driven and hybrid nowcasting paradigms, respectively. The forecast performance is evaluated using the neighborhood Critical Success Index (CSI) \cite{jolliffe2011forecast}, power spectral density (PSD) \cite{sinclair2005empirical} and Learned Perceptual Image Patch Similarity (LPIPS) \cite{zhang2018unreasonable}. All metrics are computed on column-maximum reflectivity fields to enable direct comparison with 2D baselines. The baseline configurations and metric definitions are provided in \nameref{Methods}.
In addition to reflectivity-based evaluation, we assess whether the inferred motion field is consistent with independent wind-profiler observations. The conversion from grid-cell displacements to horizontal winds and the profiler-matching procedure are described in \nameref{Methods} and Extended Data Figure~\ref{fig:wind_result}.
}

\begin{figure}[htbp]
\centering
\includegraphics[width=0.98\linewidth]{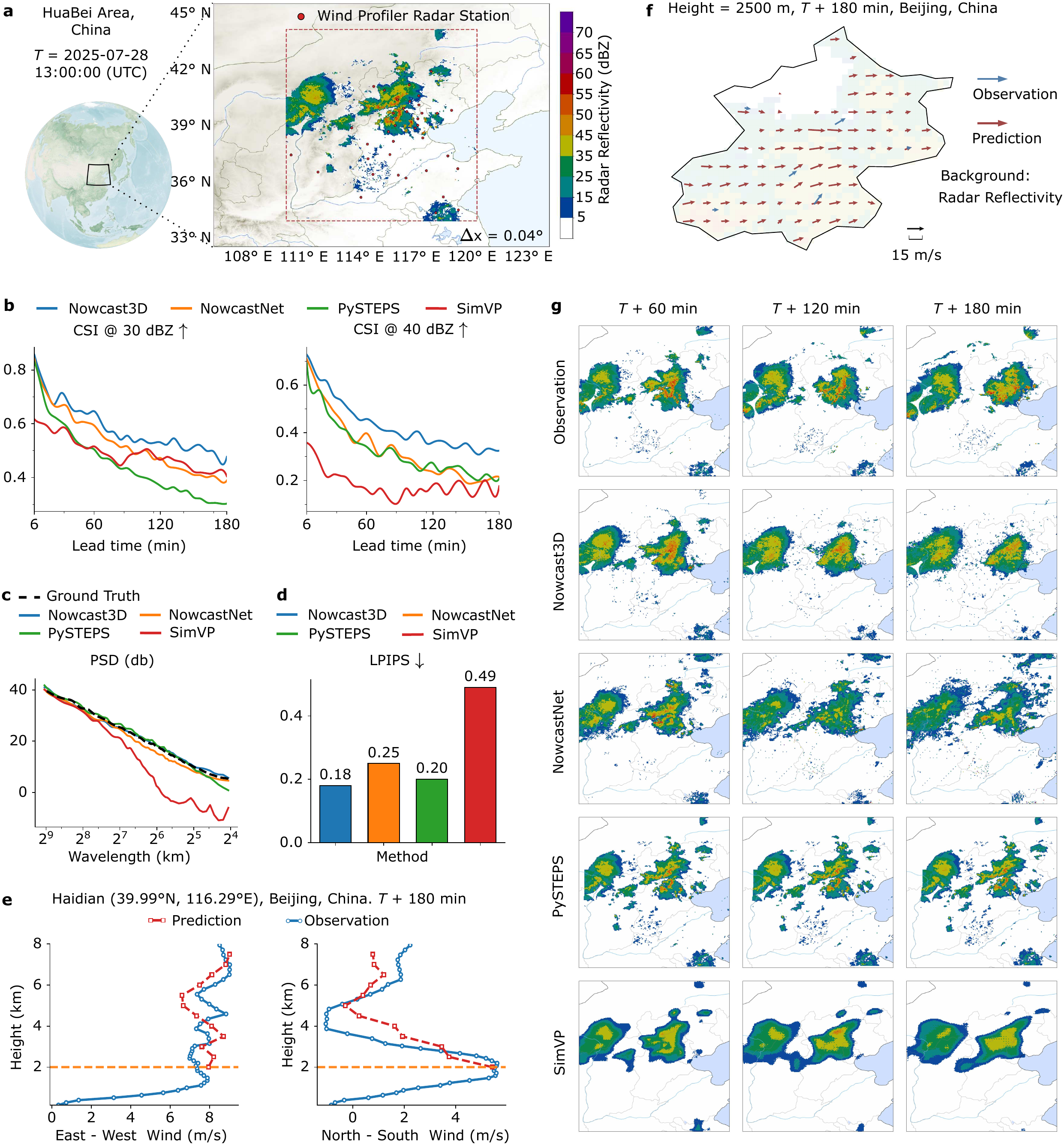}
\caption{\textbf{Nowcast3D accurately forecasts the evolution of a severe mesoscale convective system ($T$ = {13:00:00 UTC}, July 28, 2025).}
    \textbf{a}, Radar reflectivity of the convective system over North China at the forecast initialization time. The black box indicates the forecast domain, and red dots mark the locations of wind-profiler radar stations.
\textbf{b}, Quantitative comparison of forecast skills in context of CSI at 30 dBZ and 40 dBZ reflectivity thresholds over a 3-hour forecast horizon.
\textbf{c}, Analysis of spatial scale fidelity based on PSD. The forecast spectra are compared to the observed spectrum (black dashed line).
\textbf{d}, Perceptual similarity assessment based on the LPIPS metric. Lower values indicate better performance.
\textbf{e}, Vertical wind profile validation at a station located in Haidian district, Beijing. The predicted zonal (east–west) and meridional (north–south) wind components are compared with wind profiler observations at the {180-min} lead time.
\textbf{f}, Comparison of the predicted horizontal wind field with wind-profiler observations in Beijing at 2,500 m altitude ({180-min} lead time forecast).
\textbf{g}, Forecast reflectivity fields by Nowcast3D and baseline models, compared with the observed radar sequence. The frame-by-frame comparison at 6-min intervals is provided in {Supplementary Video 1}.}
\label{fig:huabei}
\end{figure}

\subsection*{Large-scale forecast at 0.04° resolution}

{We evaluate Nowcast3D on the extreme rainfall event that affected Beijing on July 28, 2025, triggering the highest-level heavy-rain warning and causing substantial societal impacts, including 32 reported fatalities. During 13{:}00--16{:}00 UTC, two convective cells evolved into a complex V-shaped echo with differential motion between its eastern and southwestern arms (Figure~\ref{fig:huabei}\textbf{a}). This case therefore tests whether a model can maintain intense echoes while tracking changing morphology.}

A comparison of the predicted and observed radar sequences highlights clear method-dependent differences (Figure~\ref{fig:huabei}\textbf{g}). 
{Among the baselines, NowcastNet preserves reflectivity amplitudes and sharp gradients relatively well, but shows substantial positional errors and misses the detailed evolution of the hazardous eastern cell. pySTEPS retains sharp echo structures, yet accumulates displacement errors under its fixed-motion extrapolation. SimVP exhibits rapid intensity loss and pronounced structural distortion.}
In contrast, Nowcast3D more closely follows the system evolution, accurately tracking the eastern cell's displacement while maintaining its internal structure and high reflectivity. 
These qualitative findings are supported by quantitative metrics. Nowcast3D achieves consistently higher neighborhood CSI scores across all lead times, indicating more accurate prediction of heavy-rainfall locations (Figure~\ref{fig:huabei}\textbf{b}). Nowcast3D also outperforms all the baselines in both PSD and LPIPS metrics, demonstrating more realistic small-scale variability and higher perceptual similarity to the observed radar fields (Figure~\ref{fig:huabei}\textbf{c}--\textbf{d}).
{The layer-wise 3D reflectivity forecasts are provided in Extended Data Figure~\ref{fig:beijing_D}.}

An important feature of our framework is its ability to infer the internal 3D wind field. To evaluate this physical component, we compare the predicted winds with independent observations from wind-profiler radars. Figure~\ref{fig:huabei}\textbf{f} shows the predicted horizontal wind field at 2{,}500\,m altitude over Beijing a {180‑min} lead time, which shows good consistency with co-located wind-profiler measurements. Figure~\ref{fig:huabei}\textbf{e} further displays the vertical wind profiles at a station located in Haidian district, Beijing. {The model reproduces consistent zonal (east--west) wind magnitudes across the vertical profile, and captures the overall meridional (north--south) vertical evolution trend.}

The inferred velocity field represents the motion of hydrometeors and is therefore most reliable in regions with substantial radar reflectivity. Its accuracy is inherently limited in echo-free areas and where low-level radar coverage is sparse. Within the precipitating volume, however, the model produces a {spatially resolved}, high-resolution wind field that complements the spatially sparse measurements by ground-based stations. The calculation and evaluation of this wind field are summarized in Extended Data Figure~\ref{fig:wind_result}, and the velocity field derived from the stream-function and potential-function decomposition is detailed in {Supplementary Note~6.3} {(see {Supplementary Figure~13})}. Another precipitation case study at a different time, together with forecast comparisons, is presented in {Supplementary Note~6.5} (see {Supplementary Figure 15} and {Supplementary Video 2}).

\begin{figure}[htbp]
\centering
\includegraphics[width=0.99\linewidth]{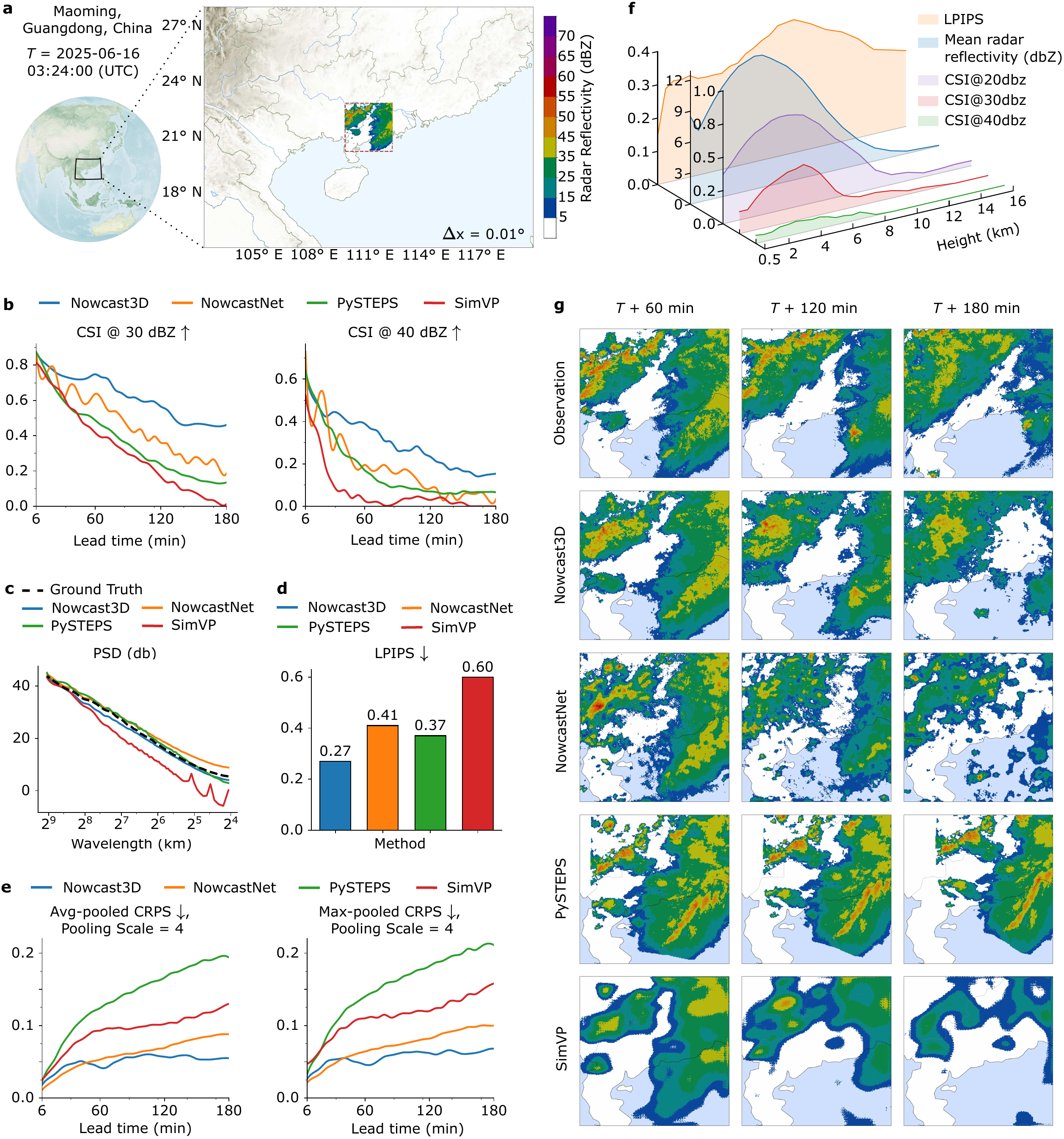}
\caption{\textbf{Generalization of Nowcast3D to high-resolution, fine-scale forecasting ($T$ = 03:24:00 UTC, June 16, 2025).}
\textbf{a}, Radar reflectivity over Maoming, Guangdong Province, China at the forecast initialization time. The analysis was conducted at 0.01° resolution.
\textbf{b}, Quantitative comparison of forecast skills in context of CSI at 30 dBZ and 40 dBZ reflectivity thresholds over a 3-hour forecast horizon.
\textbf{c}, Analysis of spatial scale fidelity based on PSD. The forecast spectra are compared to the observed spectrum (black dashed line).
\textbf{d}, Perceptual similarity assessment based on the LPIPS metric. Lower values indicate better performance.
\textbf{e}, Probabilistic forecast skill evaluated using the CRPS metric, shown as both average-pooled and max-pooled values. {Lower scores indicate better probabilistic forecast skill.}
\textbf{f}, Vertical profiles of forecast error and skill. CSI and LPIPS are plotted as a function of height, overlaid with profiles of mean observed radar reflectivity.
\textbf{g}, Forecast reflectivity fields by Nowcast3D and baseline models, compared with the observed radar sequence. The frame-by-frame comparison at 6-min intervals is provided in {Supplementary Video 3}.}
\label{fig:maoming}
\end{figure}

\subsection*{Medium-scale forecast at 0.01° resolution}

To assess our model’s ability to generalize to finer spatial scales (a key requirement for urban and operational forecasting), we evaluate its performance on forecasting an extreme precipitation event in Maoming, Guangdong Province, China on June 16, 2025. In this case, all models are operated at a 0.01° spatial resolution (Figure~\ref{fig:maoming}\textbf{a}). This higher resolution resolves faster and more intricate atmospheric processes, thereby posing a demanding test of forecast skill.

The storm evolution was dominated by two convective cells moving eastward along the coast. The western cell, initially more compact, broadened and intensified significantly as it progressed, while the eastern cell translated eastward and gradually moved out of the forecast domain. The key forecasting challenge was thus to reproduce the evolving storm morphology under rapid advection, in particular, the rapid area growth of the western cell and the simultaneous displacement of the eastern cell, which together drove large changes in the reflectivity field.

Comparing the predicted spatiotemporal evolution at this high resolution reveals the challenge of forecasting such rapidly evolving radar fields (Figure~\ref{fig:maoming}\textbf{g}). NowcastNet and SimVP, reproduce the overall eastward motion but fail to capture the expansion of the western cell, resulting in fragmented and underdeveloped echoes. pySTEPS underestimates the translation speed of the system and does not reproduce its area growth. In contrast, Nowcast3D more faithfully captures both the rapid eastward displacement and the structural expansion of the convective system.

The quantitative results support these visual assessments. Nowcast3D substantially outperforms all baseline models in neighborhood CSI, indicating more accurate localization of precipitation (Figure~\ref{fig:maoming}\textbf{b}). In addition, PSD and LPIPS evaluations show that its forecasts better reproduce the observed scale-dependent variability and exhibit higher perceptual similarity to the radar observations (Figure~\ref{fig:maoming}\textbf{c}--\textbf{d}). Because Nowcast3D produces ensembles, we also assess its probabilistic forecast skill. The Continuous Ranked Probability Score (CRPS) \cite{matheson1976scoring,hersbach2000decomposition} shows that Nowcast3D yields better-calibrated forecasts than the ensemble-based NowcastNet, indicating {improved probabilistic forecast skill} (Figure~\ref{fig:maoming}\textbf{e}).

Taking advantage of our model’s native 3D design, we further examine the vertical distribution of forecast CSI (Figure~\ref{fig:maoming}\textbf{f}). A layer-wise analysis shows that CSI is lower near the surface and at upper levels, but higher in the mid-troposphere. This pattern closely mirrors the vertical profile of mean reflectivity: echoes are strongest and most organized at mid-levels, leading to higher CSI, whereas near-surface and upper levels exhibit weaker and more intermittent echoes, naturally yielding lower CSI. The resulting vertical CSI profile is therefore closely tied to both the intrinsic storm structure and the radar sampling characteristics, rather than representing a simple ranking of forecast skill with height. In contrast, LPIPS shows much smaller inter-layer variation, indicating that the model maintains a consistent perceptual quality of forecasts across different heights. Layer-by-layer comparisons of the 3D forecasts are provided in Extended Data Figure~\ref{fig:maoming_D}.
{A case with clear vertical convective evolution is shown in {Supplementary Note~6.4} and {Supplementary Figure~14}.}
Additional precipitation case studies together with forecast comparisons are presented in {Supplementary Note~6.5} (see {Supplementary Figures 16--17} and {Supplementary Videos 4--5}).

\FloatBarrier

\begin{figure}[t!p]
\centering
\includegraphics[width=0.99\linewidth]{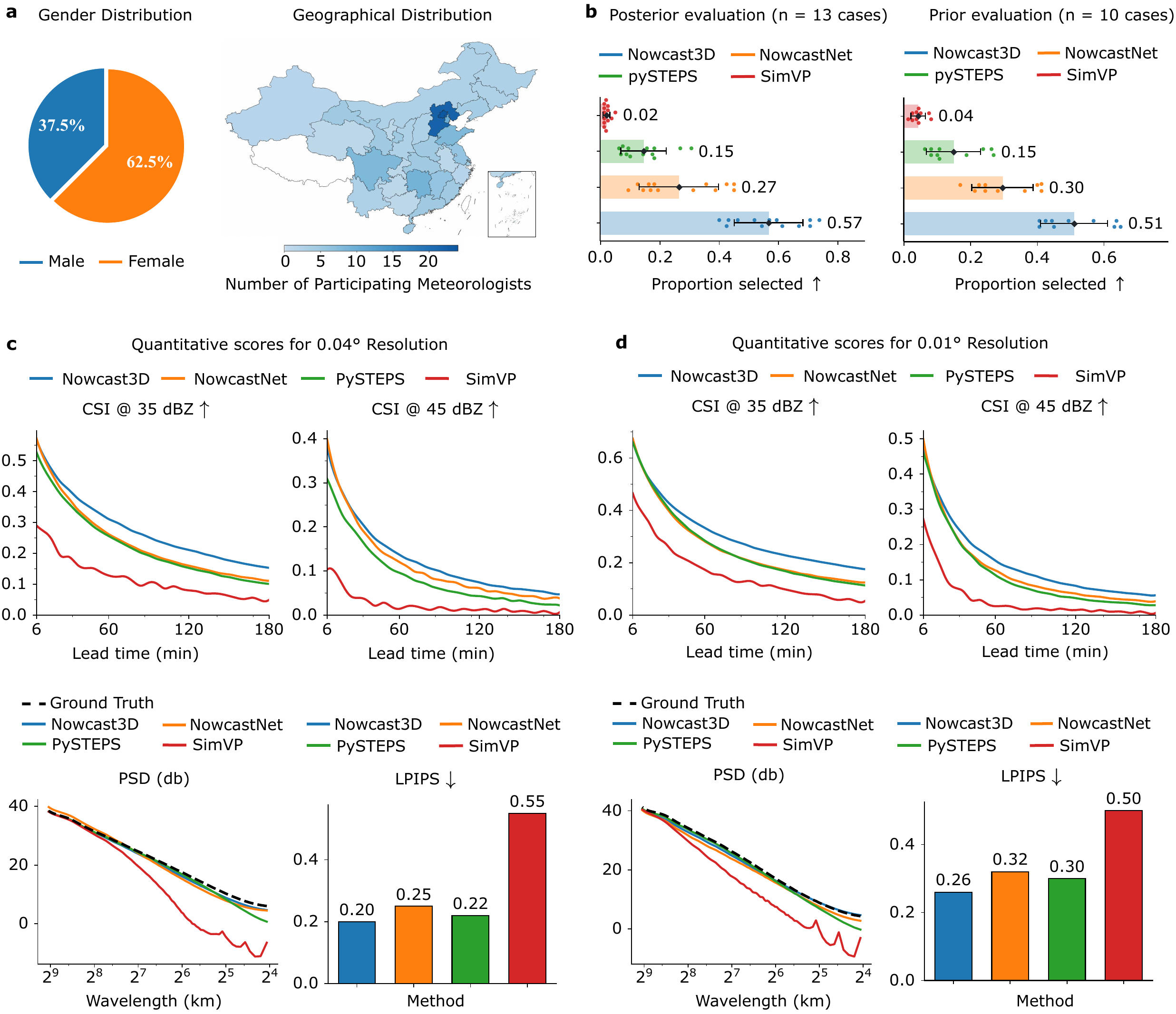}
\caption{\textbf{Aggregate quantitative evaluation of forecast skill.}
\textbf{a}, Demographics and institutional affiliations of the participating meteorologists, including gender and regional distribution.
\textbf{b}, {Model preferences in the posterior (left, \(n=13\) test cases) and prior (right, \(n=10\) test cases) evaluations. Each case was assessed by 160 meteorologists. Each point represents the proportion of meteorologists ranking a model first for one test case. Bars show the mean selection proportion across cases, with error bars indicating \(\pm1\) sample standard deviation.} Compared with other competing baseline methods, meteorologists showed a stronger preference for Nowcast3D. 
\textbf{c}, Forecast skill metrics evaluated on the 0.04° resolution test set. The top panels show CSI as a function of lead time at 35 dBZ and 45 dBZ thresholds. The bottom-left panel shows PSD of forecasts compared with the ground truth (black dashed line), and the bottom-right panel shows the mean LPIPS scores, where lower values indicate better perceptual quality.
\textbf{d}, The same set of metrics as in \textbf{c}, evaluated on the 0.01° resolution test set to assess the generalizability of different models at a finer spatial scale.
}
\label{fig:result}
\end{figure}

\subsection*{Meteorologist evaluation}

{To evaluate the operational usefulness of the forecasting models} in extreme precipitation nowcasting, we conducted an expert evaluation following the protocols established for DGMR \cite{ravuri2021skilful} and NowcastNet \cite{zhang2023skilful}. A group of senior forecasters in China were invited nationwide. Forecast animations from all models were displayed side by side, anonymized, and presented in random order to minimize bias.

A total of 160 meteorologists from the China Meteorological Administration and 30 provincial or municipal observatories participated in the {evaluation}. The demographic and regional distribution of participants is shown in Figure~\ref{fig:result}\textbf{a}. The experts cover nearly all regions of mainland China, except Tibet. Each meteorologist evaluated 23 test cases, randomly sampled from the precipitation events that occurred in China. Following the NowcastNet protocol \cite{zhang2023skilful}, we considered two complementary evaluation modes. In the post-hoc mode, forecasters ranked the model forecasts with access to the verifying radar observations. In the prior mode, they conducted ranking only using the preceding radar sequences, without access to future observations, to simulate real-time operational decision-making in the real world.

The aggregated preferences are summarized in Figure~\ref{fig:result}\textbf{b}. In the post-hoc evaluation, Nowcast3D was ranked best in 57\% of expert assessments, while, in the prior evaluation, it remained the top choice in 51\% of expert assessments. Overall, Nowcast3D received the highest preference among the participating meteorologists. These preferences are consistent with the model’s ability to represent the 3D evolution of reflectivity, enabled by its explicit treatment of physical fields and embedded dynamical constraints. In contrast, the 2D baseline models showed limitations in reproducing the full spatiotemporal structure of precipitation systems. A detailed description of the expert evaluation configuration is provided in {Supplementary Note~4}.

\subsection*{Quantitative evaluation}

To quantify the forecast skill, we employ three complementary metrics: neighborhood CSI, PSD and LPIPS. All metrics are computed over the full test corpus at two spatial scales: a provincial-scale domain at a standard 0.04° resolution and a more demanding urban domain at 0.01° resolution. This dual-scale design allows us to examine the model's performance and robustness to finer spatial details.

At 0.04° resolution, Nowcast3D consistently outperforms all baseline models across all metrics (Figure~\ref{fig:result}\textbf{c}). For both moderate (35 dBZ) and intense (45 dBZ) reflectivity thresholds, it achieves the highest neighborhood CSI throughout the 3-hour forecast horizon, indicating more accurate localization of heavy precipitation. The PSD analysis confirms that its forecasts most closely match the observed power spectrum, reflecting a more realistic representation of precipitation structures across spatial scales. This is corroborated by the LPIPS scores, where Nowcast3D attains the lowest values, indicating the highest perceptual and structural fidelity.

This advantage persists at the finer 0.01° resolution, which provides a stricter test of performance at the city scale (Figure~\ref{fig:result}\textbf{d}). Across all lead times and metrics, Nowcast3D maintains a clear margin over the competing baseline models, demonstrating that its architecture can be effectively adapted to finer spatial grids. 
Taken together, these results show that explicit modeling of 3D dynamics yields forecasts that are more accurate, structurally realistic and perceptually coherent than those produced by existing approaches. This benefit holds robustly across two different spatial scales.

{SimVP shows lower CSI at the first forecast lead time (see Figure~\ref{fig:result}\textbf{c}--\textbf{d}). Such a behavior is analyzed in {Supplementary Note~6.1} and {Supplementary Figure~11}. Additional CSI curves across a wider range of reflectivity thresholds are shown in {Supplementary Note~6.2} and {Supplementary Figure~12}.}

\begin{figure}[htbp]
\centering
\includegraphics[width=0.97\linewidth]{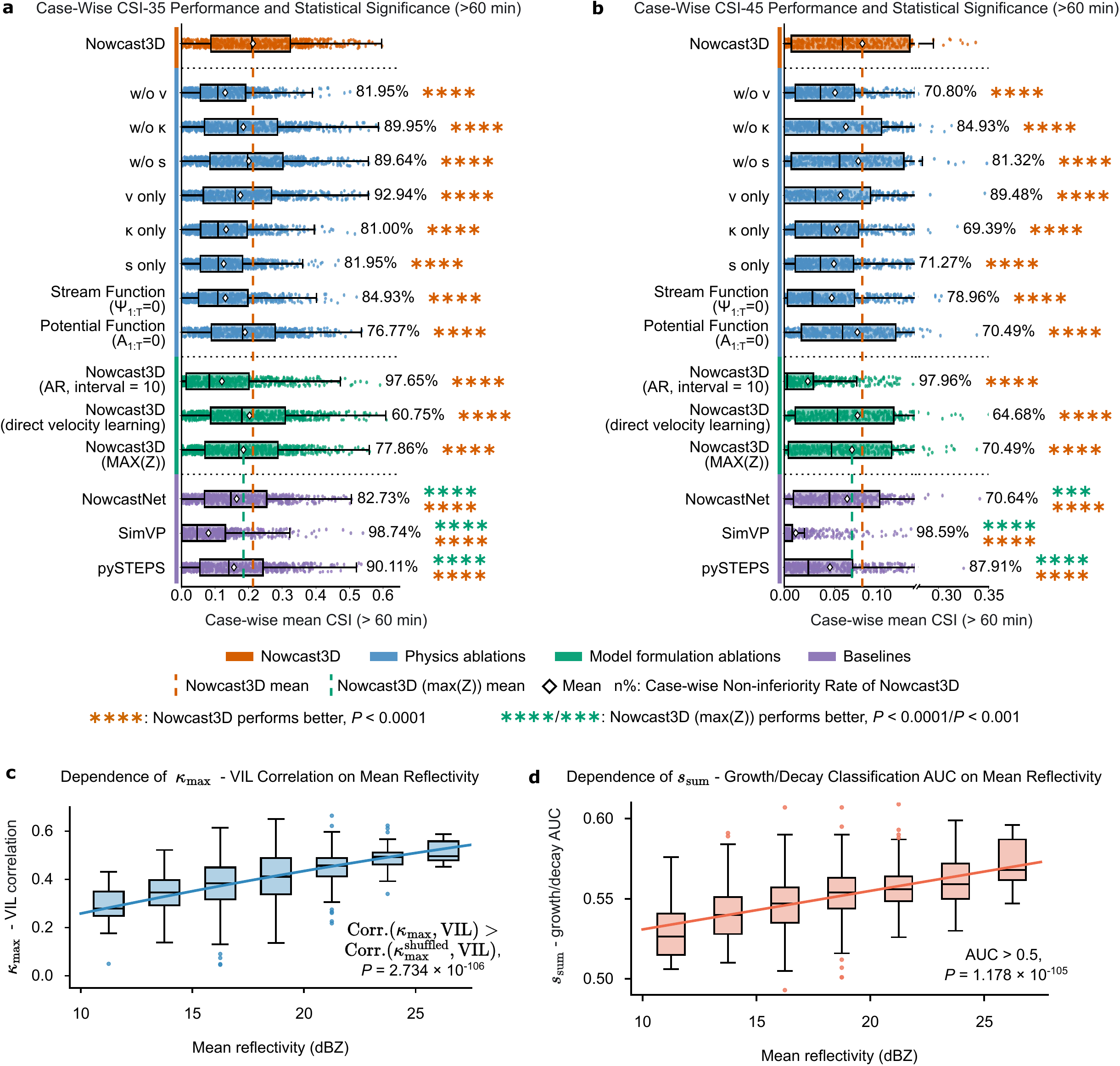}
\caption{\textbf{Case-wise ablation results and diagnostics of physical fields inferred by Nowcast3D.}
\textbf{a}, Case-wise mean neighborhood Critical Success Index (CSI) beyond 60~min at 35~dBZ across 637 test cases.
\textbf{b}, Case-wise mean neighborhood CSI beyond 60~min at 45~dBZ across the same cases.
In \textbf{a} and \textbf{b}, the colors distinguish complete Nowcast3D, physics ablations, model-formulation ablations and external baselines. The points denote individual cases, and diamonds denote the mean.
The percentages show the fraction of cases in which complete Nowcast3D matches or exceeds each alternative in case-wise CSI. The dashed lines mark the mean CSI of complete Nowcast3D and MAX$(Z)$-only Nowcast3D. The significance was assessed using one-sided Wilcoxon signed-rank tests; ****: $P<0.0001$ and ***: $P<0.001$.
\textbf{c}, Spearman correlation (Corr.) between the inferred column-maximum diffusivity magnitude, $\kappa_{{\max}}$, and vertically integrated liquid water (VIL). The shuffled reference is obtained by randomly permuting the spatial arrangement of $\kappa_{{\max}}$ (VIL unchanged). The observed correlations are significantly greater than the shuffled correlations {($P=2.734\times10^{-106}$, one-sided Wilcoxon signed-rank test)}.
\textbf{d}, Area under the receiver operating characteristic curve (AUC) for using the vertically aggregated source field $s_{\mathrm{sum}}$ to distinguish subsequent reflectivity growth from decay. An AUC above 0.5 indicates that growth columns tend to have larger inferred $s_{\mathrm{sum}}$ values than decay columns. The AUC remains greater than 0.5 {($P=1.178\times10^{-105}$, one-sided Wilcoxon signed-rank test)}.
{In all box plots, center lines denote medians, boxes span the 25th--75th percentiles, and whiskers extend to the most extreme values within 1.5 interquartile ranges of the box limits. {$P$ values were not adjusted for multiple comparisons.} Exact $P$ values for \textbf{a} and \textbf{b}, together with sample sizes for all comparisons and reflectivity bins, are provided in {Supplementary Table~2}.}
}
\label{fig:ablation_casewise}
\end{figure}

\subsection*{{Ablation study and analysis of inferred physical fields}}

{We conducted an ablation study across 637 test cases, with ablated variants of Nowcast3D. These experiments were grouped into physics ablations (considering combinations of different inferred physical fields such as velocity, diffusivity, and source) and model architecture ablations (with direct velocity learning, autoregressive coupling of deterministic and probabilistic forecast modules, or MAX$(Z)$-only input). Here, MAX$(Z)$ denotes the column-maximum reflectivity. NowcastNet, SimVP and pySTEPS were also included as baselines for comparison. Full configurations and implementation details are provided in {Supplementary Notes~5.1--5.3}.}

{We first assess the contribution of inferred physical fields to the overall performance of Nowcast3D. Case-wise ablations are evaluated over the test set using the mean neighborhood CSI of the final 20 forecast frames, thereby emphasizing the model's performance beyond the first hour. As a result, the complete Nowcast3D model yields the highest CSI distributions at both 35 and 45~dBZ (Figure~\ref{fig:ablation_casewise}\textbf{a}--\textbf{b}). Removing any physical field reduces the forecast skill, with the largest decline occurring when the velocity field is omitted. The variants retaining only one physical process also perform substantially worse than the complete model. These results indicate that the processes of advective transport, local diffusion and residual source tendencies make complementary contributions to the model's overall forecast skill.}

{The model architecture ablations further identify the contributions of different modules of Nowcast3D to its forecast skill (see Figure~\ref{fig:ablation_casewise}\textbf{a}--\textbf{b}). It can be observed that either directly learning the velocity field or autoregressive coupling of the deterministic and probabilistic modules reduces CSI compared to the complete Nowcast3D model. Restricting the input to MAX$(Z)$ also lowers the forecast skill; however, this variant still outperforms the 2D baselines. Thus, the gray-box learning scheme yields gains independent of input dimensionality, while native volumetric radar adds further skill. Across the 637 test cases, the complete Nowcast3D achieves significantly higher case-wise CSI than any ablated alternative (see the orange asterisks, $P<0.0001$, in Figure~\ref{fig:ablation_casewise}\textbf{a}--\textbf{b}), whereas the MAX$(Z)$-only variant significantly outperforms all the 2D baselines (see the green asterisks, $P<0.001$, in Figure~\ref{fig:ablation_casewise}\textbf{a}--\textbf{b}).}

{The detailed case studies and quantitative analyses of the above ablations are provided in {Supplementary Notes~5.1--5.2} and {Supplementary Figures~6--8}. The potential-based and directly-learned velocity fields are compared in {Supplementary Figure~9}. The autoregressive variants that feed Diff\mbox{-}Gen refinements back into subsequent Phy\mbox{-}Pred forecasts are evaluated in {Supplementary Note~5.3} and {Supplementary Figure~10} across different coupling intervals (e.g., one or ten forecast frames).}

{We next assess whether the inferred diffusivity and the source fields behave consistently with their assigned physical roles (see Figure~\ref{fig:ablation_casewise}\textbf{c--d}). Firstly, we compare the inferred column-maximum diffusivity magnitude ($\kappa_{{\max}}$) with the vertically integrated liquid water (VIL), where VIL is estimated from the vertical reflectivity profile and is commonly used to characterize convective-column intensity and liquid-water loading \cite{amburn1997vil}. It is seen from Figure~\ref{fig:ablation_casewise}\textbf{c} that the correlation between $\kappa_{{\max}}$ and VIL is positive and increases along with reflectivity, indicating that stronger convective columns are associated with greater effective echo dispersion. This trend is qualitatively consistent with the enhanced unresolved transport and mixing expected in intense convection \cite{shi2019key,sun2021scale}. 
Across the 637 cases, the correlations between $\kappa_{{\max}}$ and VIL are significantly greater than those between the spatially shuffled diffusivity field ($\kappa_{{\max}}^{\mathrm{shuffled}}$) and VIL ($P<0.0001$).}

{Secondly, we test whether the vertically aggregated source tendency ($s_{\mathrm{sum}}$) distinguishes growing from decaying radar columns. The growth and decay are defined by increases and decreases in the observed column-integrated reflectivity between two consecutive radar frames. An area under the receiver operating characteristic curve (AUC) above 0.5 indicates that growing columns generally have larger $s_{\mathrm{sum}}$ values than decaying columns. It is seen from Figure~\ref{fig:ablation_casewise}\textbf{d} that the case-wise AUC values are significantly greater than 0.5 ($P<0.0001$) and increase along with the mean reflectivity, showing that the inferred source field captures growth-decay tendencies, with clearer separation for stronger echoes.}

\section*{Discussion}

Precipitation nowcasting remains a central challenge in atmospheric science due to the strongly nonlinear, multiscale, and inherently non‑stationary nature of convective evolution, which contains a substantial stochastic component. Although recent data-driven advances have improved short-range guidance, the forecast skill for extreme precipitation still degrades rapidly along with lead time. Existing approaches primarily operate on 2D radar products that collapse the atmosphere into a single layer, discarding the height-dependent organization of convection and the layer-to-layer wind variations that shape storm motion and intensity through vertical motion and shear. {Consequently, 2D forecast methods neglect the physically relevant vertical structure needed for operationally useful nowcast in hazardous-weather regimes.}

Our work addresses these limitations by combining physics-informed dynamics with deep generative modeling on native 3D radar volumes. {The core idea is to establish} a physics-informed operator network that learns to infer and extrapolate key physical drivers directly from sequences of volumetric reflectivity: a 3D wind field that governs advective transport, a local field that represents stochastic dispersion, and a residual tendency associated with unresolved microphysics. 
{The forecast is advanced according to these learned operators, providing a physically structured trajectory rather than a purely black-box extrapolation.} 
{The physics-informed forecast produced by this learning core serves as the conditioning trajectory for a diffusion-based generative model}, which approximates the full distribution of future states and captures uncertainty in both storm structure and localized intensity. 
Together, these components produce forecasts that are {physically structured and probabilistically generated}, delivering the uncertainty information required for risk-aware decision-making.

The performance of this framework nonetheless depends on several factors. First, its accuracy is constrained by the quality, coverage and scanning strategy of the input radar network. The beam attenuation, ground clutter, {near-surface coverage gaps} and anomalous propagation also remain important sources of error. Second, inferring 3D flow from reflectivity alone is under-constrained, particularly in regimes with complex microphysics where phase changes can alter reflectivity without a corresponding change in the velocity field. 

These limitations point to clear directions for future work. Integrating additional observation types, such as Doppler radial velocities, geostationary satellite imagery and lightning measurements, could better constrain the inferred wind fields and improve the realism of hydrometeor evolution. Closer coupling with numerical weather prediction models, for example through data assimilation of the learned fields or by using NWP output as a physical prior, could further extend forecast skill beyond the traditional nowcasting range.

In conclusion, by unifying physics-informed dynamics with deep generative modeling on 3D radar data, this study outlines a pathway toward developing new nowcasting systems. Such models go beyond simple pattern extrapolation to learn representations of the underlying physical processes, producing forecasts that are more accurate. Further enriching this framework with multimodal observations and tighter NWP coupling has the potential to close the remaining gap toward robust, actionable forecasts of extreme precipitation events --- a capability urgently needed in practice.

\section*{Methods}\label{Methods}

\subsection*{Overview of the Nowcast3D framework}
Nowcast3D is a hybrid forecasting system that combines a physics-informed neural predictor with a diffusion-based generative model to produce probabilistic 3D radar reflectivity forecasts. As summarized in Extended Data Figure~\ref{fig:all_model}\textbf{a}, the workflow has three stages. First, raw volumetric radar observations are passed through a data-processing module that derives a 3D spatial mask of observed and unobserved regions, applies mask-guided imputation only in the unobserved regions, and normalizes reflectivity to a fixed dynamic range. Second, a physics-informed network infers underlying physical fields from the recent radar history and advances them forward in time based on an explicit advection–diffusion–source equation, thereby producing a deterministic forecast. 
Third, {a conditional diffusion model refines this physically structured forecast} into an ensemble of stochastic realizations that quantify forecast uncertainty.

In this framework, the preprocessing stage defines the input reflectivity fields $R$ used throughout the paper. The physics-informed module implements a mapping, denoted by $\boldsymbol{\Theta}$, from the observed radar history to time-varying physical fields $\boldsymbol{X}_{1:T}$ and a neural-numerical solver for the resulting evolution operator $\mathcal{F}$. The generative module implements a conditional diffusion model, denoted by $\boldsymbol{\Phi}$, that samples future trajectories conditioned on both past observations and the deterministic forecast. The following subsections describe the radar data and its preprocessing, the hybrid physics–generative formulation, the physics-informed forecasting module, the probabilistic refinement module, and the training procedure.

\subsection*{Radar datasets and preprocessing}

Our forecasts are driven solely by 3D radar reflectivity from the national radar composite operated by the China Meteorological Administration. This product provides volumetric reflectivity on 24 constant-altitude levels between 0.5 and 16~km, with native horizontal resolutions down to 0.01°. These volumes resolve the 3D structure and evolution of precipitating systems over China and form the common input to all models evaluated in this study. Details of the dataset, including the storage format, are provided in {Supplementary Note~1.1}, where {{Supplementary Figure~1} shows the height-dependent coverage curves for the study domains}.

We consider three study domains that together span diverse convective regimes and spatial resolutions. The primary training domain is located in South China (110.00°–120.24°E, 23.96°–34.20°N), where we use all eligible radar volumes recorded in 2024. Cross-regional generalization at the native 0.04° resolution is assessed on a subset of precipitating scenes over North China (110.00°–120.24°E, 33.96°–44.20°N) observed in 2025. This subset is spatially and temporally disjoint from the training domain of South China. High-resolution coastal precipitation hazards are evaluated over the Maoming region in Guangdong Province, China (109.64°–112.20°E, 20.39°–22.95°N), where pretrained models at 0.04° resolution are fine-tuned on 0.01° Maoming sequences recorded in 2024. An independent subset of Maoming precipitation events in 2025 is reserved for testing. The radar composite coverage and the three study domains are shown in Extended Data Figure~\ref{fig:radarcover}\textbf{a}. Note that Extended Data Figure~\ref{fig:radarcover}\textbf{b} illustrates how terrain blocking and the shorter range of low-elevation scans lead to a pattern of sparse, near-surface circular observations.

{Let \(R^{\mathrm{raw}}\in\mathbb{R}^{D\times H_0\times W_0}\) denote a raw three-dimensional radar reflectivity volume in dBZ, where \(D\) is the number of vertical levels and \(H_0\) and \(W_0\) are the horizontal grid dimensions of the national radar composite. Its value at vertical level \(\ell\) and horizontal grid point \((i,j)\) is denoted by \(R^{\mathrm{raw}}_{\ell ij}\).}

{Firstly, we preprocess the raw radar volumes via imputation, including missing-data identification, mask-guided low-level completion and normalization (see Extended Data Figure~\ref{fig:all_model}\textbf{b}). The missing observations are concentrated at near-surface levels owing to terrain blockage and the limited coverage of low-elevation scans (see Extended Data Figure~\ref{fig:radarcover}\textbf{b}). These unobserved grid cells must be distinguished from valid weak-echo or echo-free measurements; otherwise, they would be interpreted as artificial boundaries by the advection--diffusion solver. We therefore complete only the missing low-level regions while retaining all observed reflectivity values unchanged.}

{For each radar volume, the missing reflectivity at a low target level $\ell$ is reconstructed from the selected upper-level observations. The observed set $\mathcal{O}_{\ell}$ contains all grid points with the valid measurements at this level. At each location $(i,j)\in\mathcal{O}_{\ell}$, the selected upper-level reflectivity values are standardized to form the predictor vector $\boldsymbol{x}_{ij}$ and paired with the target-level reflectivity $R^{\mathrm{raw}}_{\ell ij}$. A separate ridge-regression model is fitted for each radar volume and target level:}
\begin{equation}
{
\boldsymbol{\theta}_{\ell}^{\star}
=
\arg\min_{\boldsymbol{\theta}_{\ell}}
\sum_{(i,j)\in\mathcal{O}_{\ell}}
\left(
R^{\mathrm{raw}}_{\ell ij}
-
\boldsymbol{\theta}_{\ell}^{\top}\boldsymbol{x}_{ij}
\right)^2
+
\alpha\lVert\boldsymbol{\theta}_{\ell}\rVert_2^2.
}
\end{equation}
{where $\boldsymbol{\theta}_{\ell}$ denotes the regression coefficient vector for level $\ell$ (with $\boldsymbol{\theta}_{\ell}^{\star}$ being its fitted value) and $\alpha$ controls the strength of ridge regularization. The fitted model yields the two-dimensional initial estimate $\widehat{R}_{\ell}$, with $\widehat{R}_{\ell ij}=(\boldsymbol{\theta}_{\ell}^{\star})^{\top}\boldsymbol{x}_{ij}$.}

{Because the regression estimate does not explicitly enforce horizontal continuity, $\widehat{R}_{\ell}$ may contain discontinuities near the boundaries of missing regions. We therefore refine the missing region selected for completion ($\mathcal{U}_{\ell}$) using an edge-preserving guided filter, with the mean reflectivity across the selected upper levels ($\bar{R}^{\mathrm{high}}$) as the guide, followed by level-specific intensity calibration:}
\begin{equation}
\label{eq:imputation_refinement}
{
R^{\mathrm{imp}}_{\ell ij}
=
\begin{cases}
\displaystyle
\mathcal{C}_{\ell}
\left[
\mathrm{GF}_{\bar{R}^{\mathrm{high}}}
\left(
\widehat{R}_{\ell}
\right)
\right]_{ij},
&
(i,j)\in\mathcal{U}_{\ell},
\\[10pt]
R^{\mathrm{raw}}_{\ell ij},
&
(i,j)\in\mathcal{O}_{\ell}.
\end{cases}
}
\end{equation}
{where $R^{\mathrm{imp}}_{\ell ij}$ denotes the completed reflectivity at level $\ell$. The operator $\mathrm{GF}_{\bar{R}^{\mathrm{high}}}$ filters the regression estimate using $\bar{R}^{\mathrm{high}}$ as the structural guide, and $\mathcal{C}_{\ell}$ calibrates the completed intensities at that level. Only the missing grid points ($(i,j)\in\mathcal{U}_{\ell}$) are updated, whereas the values at the observed locations ($(i,j)\in\mathcal{O}_{\ell}$) remain unchanged.}

{The completed reflectivity volumes are subsequently normalized to a fixed dynamic range. Extended Data Figure~\ref{fig:radarcover}\textbf{c}--\textbf{d} shows the representative low-level fields before and after completion at 500 and 1{,}000~m over South China and Maoming. Extended Data Figure~\ref{fig:radarcover}\textbf{e} provides a height-resolved cross-section over North China, illustrating the completed near-surface field relative to the observed vertical echo structure. The missing-data definition, implementation parameters and limitations are discussed in {Supplementary Notes~1.2--1.3}.}

\subsection*{Hybrid physics–generative formulation}

Let \(\{R_t\}_{t=-T_0}^{T}\) denote the imputed and normalized 3D radar reflectivity sequence, spatially cropped to a target domain with horizontal grid dimensions \(H \times W\). Each \(R_t \in \mathbb{R}^{D \times H \times W}\). The past observations are written as $R_{-T_0:0}\in\mathbb{R}^{(T_0+1)\times D\times H\times W}$. Given $R_{-T_0:0}$, our goal is to forecast the future sequence $R_{1:T}\in\mathbb{R}^{T\times D\times H\times W}$ using a unified two-stage framework that first produces a physics-based deterministic forecast and then refines it by a probabilistic ensemble model.

In the physics-informed learning stage, a neural network $\boldsymbol{\Theta}$ maps the historical radar sequence to a set of time-varying physical fields,
$\boldsymbol{X}_{1:T}=\boldsymbol{\Theta}(R_{-T_0:0})$, where $\boldsymbol{X}_{1:T}\in\mathbb{R}^{T\times C\times D\times H\times W}$ and $C$ is the number of inferred quantities that govern the evolution of reflectivity. These fields act as coefficients in a time-dependent evolution equation,
$\partial_t R=\mathcal{F}(R,\boldsymbol{X})$, where $\mathcal{F}$ represents the physical dynamics controlled by $\boldsymbol{X}$. Starting from the initial condition $R(0)=R_0$ (the last observed volume), integrating this equation forward in time yields a deterministic forecast $R^{\mathrm{det}}_{1:T}$.

In the probabilistic forecasting stage, {a conditional diffusion model generates an ensemble of forecasts around the physically structured trajectory produced by the deterministic core.} The generative network $\boldsymbol{\Phi}$ is conditioned on the context
$\boldsymbol{c}=\{R_{-T_0:0},R^{\mathrm{det}}_{1:T}\}$, which combines the past observations with the deterministic prediction. Ensemble members are obtained by sampling the learned conditional distribution with different noise realizations $\epsilon^{(k)}$, $k=1,\dots,K$, thereby representing stochastic variability and forecast uncertainty.

The aforementioned steps can be written compactly as follows:
\begin{subequations}\label{eq:framework} 
\renewcommand{\theequation}{\theparentequation.\arabic{equation}} 
\begin{align} 
\boldsymbol{X}_{1:T} &= \boldsymbol{\Theta}\big(R_{-T_0:0}\big),\quad X_{1:T}\in\mathbb{R}^{T\times C\times D\times H\times W}, &&\text{(physical fields)} \label{eq:framework-phys}\\[4pt] 
R^{\mathrm{det}}_{1:T} &= \mathrm{Solve}\Bigl( \frac{\partial R}{\partial t}=\mathcal{F}(R,\boldsymbol{X}),\;R(0)=R_0\Bigr), && \text{(deterministic forecast)} \label{eq:framework-det}\\[4pt] 
\boldsymbol{c} &= \{\,R_{-T_0:0},\;R^{\mathrm{det}}_{1:T}\,\}, && \text{(conditioning context)} \label{eq:framework-cond}\\[4pt] 
R^{(k)}_{1:T} &= \mathrm{Sample}_{\boldsymbol{\Phi}}\big(\epsilon^{(k)},\,\boldsymbol{c}\big), \;k=1,\dots,K, && \text{(sampling)} \label{eq:framework-sample}\\[4pt] 
R^{\mathrm{ens}}_{1:T} &= \{\,R^{(k)}_{1:T}\,\}_{k=1}^K. && \text{(ensemble forecast)} 
\label{eq:framework-ens} 
\end{align} 
\end{subequations}
Nowcast3D implements these formulations with a shared backbone that realizes $\boldsymbol{\Theta}$ and a conditional diffusion network that realizes $\boldsymbol{\Phi}$, as described in the following.

\subsection*{Physics-informed forecasting module}

\paragraph{Network architecture.}
The physics-informed forecasting module (aka, the Phy-Pred Network shown in Extended Data Figure~\ref{fig:all_model}\textbf{c}) implements the mapping $\boldsymbol{\Theta}$ in Eq.~(\ref{eq:framework-phys}). An encoder first extracts multi-scale features from the preprocessed radar history $R_{-T_0:0}$. These features are processed by two complementary branches: a U-Net branch \cite{ronneberger2015u} that captures local and fine-scale structures, and a Transformer branch \cite{vaswani2017attention} that represents long-range spatial and temporal dependencies. Their outputs are fused into a common representation, which is then decoded into time-varying physical fields corresponding to advective motion, anisotropic diffusion and intensity tendency. These fields parameterize the evolution operator $\mathcal{F}$, and a numerical solver advances the last observed radar volume in time to obtain the deterministic forecast $R^{\mathrm{det}}_{1:T}$.

\paragraph{Physical dynamics.}
We model the evolution of radar reflectivity in continuous time and space over a fixed forecast horizon.
The deterministic forecast $R^{\mathrm{det}}_{1:T}$ is obtained by numerically integrating a partial differential equation. We specify the dynamics $\mathcal{F}(R,X)$ as an advection–diffusion–source equation, given by
\begin{equation}
\frac{\partial R}{\partial t}
=
\underbrace{-\boldsymbol{v}\cdot\nabla R}_{\text{advection}}
\;+\;
\underbrace{\nabla\cdot(\boldsymbol{\kappa}\nabla R)}_{\text{diffusion}}
\;+\;
\underbrace{s}_{\text{source}}.
\label{eq:ns}
\end{equation}
Here, the reflectivity field $R$ evolves under three processes: advective transport by a 3D velocity field $\boldsymbol{v}$; local anisotropic diffusion governed by a diffusivity tensor $\boldsymbol{\kappa}$; and a residual source term $s$ that summarizes unresolved processes, including microphysical growth and decay. The time-varying fields $(\boldsymbol{v},\boldsymbol{\kappa},s)$ over the forecast horizon constitute the physical fields $X_{1:T}$ and are inferred from the input radar sequence $R_{-T_0:0}$ by the physics module $\boldsymbol{\Theta}$.

\paragraph{Neural parameterization.}
The time-varying physical fields
$X_{1:T} = (\boldsymbol{v}_{1:T}, \boldsymbol{\kappa}_{1:T}, s_{1:T})$
required in Eq.~\eqref{eq:ns} are produced by a shared encoder–decoder backbone followed by task-specific heads. We denote the backbone (encoder, U-Net branch, Transformer branch and fusion) by $\mathrm{Backbone}(\cdot)$. Its output is passed to three dedicated decoders that specialize in motion, diffusion and source components, expressed as follows:
\begin{subequations}\label{eq:neural_mapping}
  \renewcommand{\theequation}{\theparentequation.\arabic{equation}}
  \begin{align}
  (\varphi_{1:T}, \boldsymbol{\psi}_{1:T})
    &= \mathrm{Decoder}_{\mathrm{Motion}}\!\big(\mathrm{Backbone}(R_{-T_0:0})\big),
    \label{eq:neural_mapping-1}\\[4pt]
  \boldsymbol{\kappa}_{1:T}
    &= \mathrm{Decoder}_{\mathrm{AnisoBrown}}\!\big(\mathrm{Backbone}(R_{-T_0:0})\big),
    \label{eq:neural_mapping-2}\\[4pt]
  s_{1:T}
    &= \mathrm{Decoder}_{\mathrm{Intensity}}\!\big(\mathrm{Backbone}(R_{-T_0:0})\big),
    \label{eq:neural_mapping-3}
  \end{align}
\end{subequations}
where the full velocity field is reconstructed from a scalar potential $\varphi$ and a vector potential $\boldsymbol{\psi}$ via
$\boldsymbol{v}_{1:T} = \nabla \varphi_{1:T} + \nabla \times \boldsymbol{\psi}_{1:T}.$
The backbone parameters are shared across all outputs, whereas the three decoders have separate parameters and specialize in inferring the dynamics of motion, diffusion and source terms, respectively. Additional architecture details are given in {Supplementary Figure 2}.

\paragraph{Numerical extrapolation.}
Given the inferred fields $(\boldsymbol{v}_{1:T}, \boldsymbol{\kappa}_{1:T}, s_{1:T})$, we advance the radar field $R_t$ forward in time by numerically integrating Eq.~\eqref{eq:ns}. We use an operator-splitting scheme that decomposes each time step into successive advection, diffusion and source updates. The advection operator that reads
\begin{equation}
\mathcal{A}_{\Delta t}[R_t](x) = R_t\bigl(x - \boldsymbol{v}_t(x)\,\Delta t\bigr)
\end{equation}
is implemented using a semi-Lagrangian method \cite{staniforth1991semi,zhang2023skilful}, which remains stable for large Courant numbers and is well suited to long lead times. To implement anisotropic diffusion, we exploit the link between diffusion and random motion \cite{einstein1905molekularkinetischen,chapman1990mathematical,zhang2025monte} and define the diffusion operator as
\begin{equation}
\mathcal{D}_{\Delta t}[R_t](x) = \mathbb{E}_{\eta \sim \mathcal{N}\bigl(0,\,2\boldsymbol{\kappa}_t \Delta t\bigr)}
\left[ R_t(x + \eta) \right],
\end{equation}
where the expectation is taken over Gaussian displacements with covariance determined by $\boldsymbol{\kappa}_t$. The source term is applied as an additive correction.
{Because scattering contributions are additive, averaging can alternatively be performed in linear reflectivity space. Additional discussion is provided in {Supplementary Note~2.1}.} 

Composing these operators yields the one-step update expressed as
\begin{equation}
R_{t+1} \approx \mathcal{D}_{\Delta t}\big[\mathcal{A}_{\Delta t}[R_t]\big] + s_t\,\Delta t.
\label{eq:operator_update}
\end{equation}
In practice, the expectation in the diffusion step is approximated by a Monte Carlo average over $M$ samples. For a voxel at location $x$, the numerical update becomes
\begin{equation}
\label{eq:final-update}
R_{t+1}(x)
\approx
\frac{1}{M}\sum_{m=1}^{M}
R_{t}\!\big( x - \boldsymbol{v}_t(x)\,\Delta t + \eta^{(m)}_t(x) \big)
+ s_t(x)\,\Delta t,
\end{equation}
where each stochastic displacement $\eta^{(m)}_t(x)$ is independently sampled from
$\mathcal{N}\bigl(0,\,2\boldsymbol{\kappa}_t(x)\Delta t\bigr)$.

{For all deterministic forecasts, one solver step corresponds to a 6-min interval between consecutive radar volumes ($\Delta t=6$~min). No adaptive time stepping or temporal sub-stepping is used. Each update comprises one semi-Lagrangian advection step, one Monte Carlo Brownian diffusion step with $M=8$ samples, and one additive source update. We employ the same setting of operator-splitting sequence, time step and Monte Carlo sample count for the $0.04^\circ$ and $0.01^\circ$ experiments.}

{
\paragraph{Calculation and evaluation of inferred motion fields.}
To assess the consistency of the velocity field $\boldsymbol{v}$ inferred by Nowcast3D with independent atmospheric observations, we compare it with the measurement from wind-profiler radars (see Extended Data Figure~\ref{fig:wind_result}). The comparison includes two steps. First, the model output, expressed as grid-cell displacement over one forecast interval, is converted into equivalent horizontal velocity components in m/s. Let $\lambda$ and $\phi$ denote longitude and latitude, respectively, and $\Delta\theta$ the angular grid spacing. If $L_x$ and $L_y$ are the inferred east-west and north-south displacements, measured in grid cells over the interval $\Delta t$, the corresponding velocity components are
}
\begin{equation}
\label{eq:velocity_conversion}
{
u
=
\frac{L_x\,\Delta\theta\,\pi r_x\cos(\phi)}
{180\,\Delta t},
\qquad
v
=
\frac{L_y\,\Delta\theta\,\pi r_y}
{180\,\Delta t}.
}
\end{equation}
{where $r_x$ and $r_y$ are the local metric radii used to convert angular displacements into physical distances in the east-west and north-south directions. }

{Second, each wind-profiler observation is matched to the inferred velocity field in space, time, and altitude. Because profiler sites rarely coincide with velocity grid points, we define a cylindrical neighborhood around each observation. Its horizontal and vertical extents are determined by the search radius and altitude tolerance (Extended Data Figure~\ref{fig:wind_result}\textbf{b}--\textbf{c}). The velocity vectors within this neighborhood at the corresponding forecast time are averaged to obtain the matched estimate.}

{The matched pairs are used to calculate the mean absolute error (MAE) and the bias of the east-west and north-south velocity components over the forecast horizon (Extended Data Figure~\ref{fig:wind_result}\textbf{d}--\textbf{e}). The MAE remains relatively stable, with only a weak increase at longer lead times for both components. The east-west bias changes sign over time and remains comparatively small. In contrast, the north-south component shows a predominantly negative bias that becomes more pronounced at several later lead times. These results show that the inferred velocity maintains a broadly consistent error magnitude over the forecast horizon. Further analysis of the velocity field reconstructed from the scalar-potential and vector stream-function parameterization is provided in {Supplementary Note~6.3} and {Supplementary Figure~13}.}

\subsection*{Probabilistic refinement module}

\paragraph{Network architecture.}
The diffusion-based generative network (aka, the Diff-Gen Network depicted in Extended Data Figure~\ref{fig:all_model}\textbf{d}) implements the mapping $\boldsymbol{\Phi}$ in Eq.~(\ref{eq:framework-sample}). We adopt a conditional generative architecture that explicitly incorporates historical information \cite{yu2024diffcast}: given the conditioning information $c=\{R_{-T_0:0},R^{\mathrm{det}}_{1:T}\}$, a ContextNet first encodes this information into a compact representation that summarizes both the recent observations and the trajectory of physics-informed forecasts. Conditional diffusion modules then take this context together with independent noise realizations and iteratively refine them into stochastic forecast trajectories. One branch focuses on the large-scale reflectivity pattern, while a residual diffusion branch learns local intensity corrections relative to the deterministic forecast. Different noise inputs produce a set of samples $\{R^{(k)}_{1:T}\}_{k=1}^K$, which together form the ensemble forecast $R^{\mathrm{ens}}_{1:T}$ used for probabilistic evaluation and downstream decision-making (see {Supplementary Figure~3} for architecture details).

Because the physics-informed learning module already advances the full 3D reflectivity field through explicit dynamical operators, the role of $\boldsymbol{\Phi}$ is to provide a stochastic refinement around this baseline state. To reduce training complexity and computational cost, we implement this refinement in the 2D space, operating on full time sequences of radar cross-sections rather than on complete 3D volumes. The same trained network can subsequently be applied either to column-maximum reflectivity sequences to generate 2D probabilistic forecasts, or independently to sequences from multiple height levels, whose outputs can be stacked to reconstruct a 3D ensemble of reflectivity volumes. Further construction and sampling details are provided in {Supplementary Note~2.2}.

\paragraph{Ensemble construction.}
Operationally, the probabilistic refinement module implements the sampling step
$R^{(k)}_{1:T} = \mathrm{Sample}_{\boldsymbol{\Phi}}(\epsilon^{(k)}, c)$
used to construct the ensemble forecast. For each noise realization $\epsilon^{(k)}$, the two diffusion branches produce a pair of correlated outputs: a full-field structure $S_{\mathrm{struct}}^{(k)}$ and a residual field $S_{\mathrm{res}}^{(k)}$ defined relative to the deterministic forecast. This pair of outputs supports three types of ensemble members constructed from each sample, namely,
\begin{subequations}\label{eq:ensemble_construction}
  \renewcommand{\theequation}{\theparentequation.\arabic{equation}}
  \begin{align}
  S^{(1)} &= S_{\mathrm{struct}},
           && \text{(full-field forecast)}
           \label{eq:ensemble_construction-1}\\[4pt]
  S^{(2)} &= R^{\mathrm{det}}_{1:T} + S_{\mathrm{res}},
           && \text{(physics-corrected forecast)}
           \label{eq:ensemble_construction-2}\\[4pt]
  S^{(3)} &= S_{\mathrm{struct}} + \alpha S_{\mathrm{res}},
           && \text{(hybrid forecast)}
           \label{eq:ensemble_construction-3}
  \end{align}
\end{subequations}
where $\alpha\in[0,1]$ is a mixing coefficient. Generating $K$ independent noise realizations yields $K$ pairs $\{(S_{\mathrm{struct}}^{(k)}, S_{\mathrm{res}}^{(k)})\}_{k=1}^K$, from which we form the ensemble:
\begin{equation}
\boldsymbol{R}^{\mathrm{ens}}_{1:T}
=
\big\{
S^{(i)}_{1:T}(k) \;:\; i\in\{1,2,3\},\; k=1,\dots,K
\big\},
\end{equation}
where \(S^{(i)}_{1:T}(k)\) denotes the \(i\)-th construction in Eq.~\eqref{eq:ensemble_construction} for sample \(k\). In the hybrid formulation, all these members are collectively denoted by \(R^{\mathrm{ens}}_{1:T}\) and are used for probabilistic evaluation. {All ensemble members are conditioned on the same physics-informed forecast. Generated perturbations are not fed back into the physical solver, so early-time structural uncertainty is not explicitly propagated through the physical dynamics. In rapidly evolving convective systems, this may affect the reliability of ensemble spread at longer lead times.} {The effect of introducing such feedback is examined in {Supplementary Note~5.3}. Using the existing model, we found no consistent improvement in long-lead CSI with autoregressive feedback.}
{{Supplementary Note~2.3} specifies how the structure-only, residual-corrected and blended members are grouped into the final ensemble. {Supplementary Figure~4} shows an example of the ensemble forecasts and the associated confidence maps, while the member-type ablations and quantitative comparisons are provided in {Supplementary Note~5.2} and {Supplementary Figures~7--8}.}

\subsection*{Training}

\paragraph{Training protocol.}
The model training proceeds in two stages that mirror the hybrid formulation described above. In the first stage, we train the physics-informed module $\boldsymbol{\Theta}$ to produce a deterministic evolution of the radar reflectivity field. In the second stage, we freeze $\boldsymbol{\Theta}$ and train the probabilistic refinement module $\boldsymbol{\Phi}$ to learn a conditional distribution of futures given both the past observations and the deterministic forecast. {The key training and inference settings, including sample counts, epoch budgets, hardware and measured runtimes, are consolidated in {Supplementary Table 1}.}

\paragraph{Training the physics-informed module.}
In the first stage, we optimize all parameters of the physics-informed module $\boldsymbol{\Theta}$, including the shared backbone and the three task-specific decoders. The objective $\mathcal{L}_{\mathrm{phys}}$ is a composite loss that supervises the intermediate states of the operator sequence as well as the final forecast. At each forecast step $t$, we penalize discrepancies between (i) the purely advected field, (ii) the advected–diffused field and (iii) the final source-corrected forecast and the corresponding ground-truth radar volume at time $t+1$. The total loss is a weighted sum of $\ell_1$ norms accumulated over the forecast horizon, namely,
\begin{equation}\label{eq:L_phys}
\mathcal{L}_{\mathrm{phys}} =
\sum_{t=0}^{T-1}
\Big(
\lambda_{\text{adv}}\mathcal{L}_{\text{adv}}^{(t)}
+\lambda_{\text{diff}}\mathcal{L}_{\text{diff}}^{(t)}
+\mathcal{L}_{\text{pred}}^{(t)}
\Big),
\end{equation}
where 
\begin{subequations}\label{eq:phys_losses}
  \renewcommand{\theequation}{\theparentequation.\arabic{equation}}
  \begin{align}
  \mathcal{L}_{\text{adv}}^{(t)}
    &= \big\|\mathcal{A}_{\Delta t}[R_t] - R_{t+1}^{\text{truth}}\big\|_1,
    \label{eq:phys_losses-adv}\\[4pt]
  \mathcal{L}_{\text{diff}}^{(t)}
    &= \big\|\mathcal{D}_{\Delta t}\big[\mathcal{A}_{\Delta t}[R_t]\big]
          - R_{t+1}^{\text{truth}}\big\|_1,
    \label{eq:phys_losses-diff}\\[4pt]
  \mathcal{L}_{\text{pred}}^{(t)}
    &= \big\|R_{t+1} - R_{t+1}^{\text{truth}}\big\|_1.
    \label{eq:phys_losses-pred}
  \end{align}
\end{subequations}
Here, $R_t$ and $R_{t+1}$ denote the model forecasts at times $t$ and $t+1$, respectively, and $R_{t+1}^{\mathrm{truth}}$ is the corresponding observed radar volume. The weights $\lambda_{\mathrm{adv}}$ and $\lambda_{\mathrm{diff}}$ control the contributions of the intermediate advection and diffusion losses. {In all experiments, the three supervised terms are weighted equally, with $\lambda_{\mathrm{adv}}=1$, $\lambda_{\mathrm{diff}}=1$ and unit weight assigned to the final prediction loss.} These intermediate objectives encourage the network to learn process-specific advection and diffusion representations in addition to accurate end-of-step forecasts. {Further training details are provided in {Supplementary Note~3.1}. {Supplementary Figure~5} compares the training-loss convergence and corresponding high-threshold CSI performance under different loss configurations.}

\paragraph{Training the probabilistic refinement module.}
In the second stage, we train the probabilistic refinement module $\boldsymbol{\Phi}$ while keeping $\boldsymbol{\Theta}$ fixed. Its dual-branch architecture is designed to produce the structured field $S_{\mathrm{struct}}$ and residual field $S_{\mathrm{res}}$ used in the ensemble construction in Eq.~\eqref{eq:ensemble_construction}: the structure branch learns a distribution over full future reflectivity fields, whereas the residual branch learns a distribution over corrections to the deterministic forecast.

We train both branches on 2D radar fields rather than on full 3D radar volumes. For each spatiotemporal window, we treat either a single height level or the column-maximum projection as a 2D sample, yielding an augmented training set $R^{\mathrm{2D}}$ with corresponding historical, deterministic and truth sequences $R^{\mathrm{2D}}_{\text{hist}}$, $R^{\mathrm{2D,det}}_{1:T}$ and $R^{\mathrm{2D,truth}}_{1:T}$. 

Each branch is trained as a conditional diffusion model using a shared noise schedule and the $v$-parameterization \cite{salimansprogressive}. For notational simplicity, we write $R_{\text{truth}}$ for $R^{\mathrm{2D,truth}}_{1:T}$ and
$r_{\text{truth}} = R_{\text{truth}} - R^{\mathrm{2D,det}}_{1:T}$ for the corresponding residual composites. The 2D conditioning context is given by 
$c = \bigl(R^{\mathrm{2D}}_{-T_0:0},\,R^{\mathrm{2D,det}}_{1:T}\bigr)$.
Here, $\tau$ indexes the diffusion step. We sample $\tau\sim\mathrm{Unif}\{0,\dots,N-1\}$ and $\epsilon\sim\mathcal{N}(0,\mathbf{I})$, where $N$ is the total number of diffusion steps. Let $\{\beta_\tau\}_{\tau=0}^{N-1}$ be the shared noise schedule, $\alpha_\tau = 1-\beta_\tau$, and $\bar{\alpha}_\tau = \prod_{s=0}^{\tau}\alpha_s$. 
For each branch, a dedicated context encoder processes $c$ and produces a hierarchy of multi-scale feature maps used to condition the denoiser; we denote these encoded features by
$h_{\text{struct}} = E_{\phi_{\text{struct}}}(c)$ and $h_{\text{res}} = E_{\phi_{\text{res}}}(c)$.
The structure and residual branches are trained with the following diffusion losses:
\begin{subequations}\label{eq:diffusion_losses}
  \renewcommand{\theequation}{\theparentequation.\arabic{equation}}
  \begin{align}
  \mathcal{L}_{\text{struct}}
    &= \mathbb{E}_{\tau,\,R_{\text{truth}},\,\epsilon} \Big[
       \big\|
          \underbrace{
            \sqrt{\bar{\alpha}_\tau}\,\epsilon
            - \sqrt{1-\bar{\alpha}_\tau}\,R_{\text{truth}}
          }_{v_\tau^{\text{struct}}}
          -
          v_{\theta_{\text{struct}}}\big(
            \underbrace{
              \sqrt{\bar{\alpha}_\tau}\,R_{\text{truth}}
              + \sqrt{1-\bar{\alpha}_\tau}\,\epsilon
            }_{x_\tau^{\text{struct}}},
            \tau,\,h_{\text{struct}}
          \big)
       \big\|_2^2 \Big],
    \label{eq:diffusion_losses-struct}\\[4pt]
  \mathcal{L}_{\text{res}}
    &= \mathbb{E}_{\tau,\,r_{\text{truth}},\,\epsilon} \Big[
       \big\|
          \underbrace{
            \sqrt{\bar{\alpha}_\tau}\,\epsilon
            - \sqrt{1-\bar{\alpha}_\tau}\,r_{\text{truth}}
          }_{v_\tau^{\text{res}}}
          -
          v_{\theta_{\text{res}}}\big(
            \underbrace{
              \sqrt{\bar{\alpha}_\tau}\,r_{\text{truth}}
              + \sqrt{1-\bar{\alpha}_\tau}\,\epsilon
            }_{x_\tau^{\text{res}}},
            \tau,\,h_{\text{res}}
          \big)
       \big\|_2^2 \Big],
    \label{eq:diffusion_losses-res}
  \end{align}
\end{subequations}
where $\epsilon\sim\mathcal{N}(0,\mathbf{I})$ emphasizes the Gaussian noise, $\bar{\alpha}_\tau$ is the cumulative product of the noise schedule, and $v_{\theta_{\text{struct}}}$ and $v_{\theta_{\text{res}}}$ denote the predicted $v$-parameters for the structure and residual branches, respectively. {Both diffusion branches use \(N=1000\) training diffusion steps, a sigmoid noise schedule and a deterministic DDIM sampler with 250 reverse steps at inference. The structure and residual branches are trained with the same optimizer settings but separate branch-specific targets.} Further details are provided in {Supplementary Note~3.2}.

\paragraph{Forecast configurations and transfer learning.}
We evaluate the model performance at two spatial resolutions: \(0.04^\circ\) (coarse) and \(0.01^\circ\) (fine). At both resolutions, we use \(256\times256\)-pixel input patches. For the \(0.04^\circ\) experiments, this corresponds to a \(10.24^\circ\times10.24^\circ\) domain that is representative of province-scale regions. For the \(0.01^\circ\) experiments, it corresponds to a \(2.56^\circ\times2.56^\circ\) domain tailored to high-resolution, city-scale hazards.

{The fine-resolution experiment follows a transfer-learning setup: models are pretrained on South China sequences at $0.04^\circ$ resolution and then fine-tuned on Maoming sequences at $0.01^\circ$ resolution (both recorded in 2024), with independent Maoming events in 2025 reserved for testing. The network retains the same $256\times256$ tensor dimensions and numerical solver, including the radar-frame time step and Monte Carlo sample count. Resolution is represented by the horizontal coordinates which are transformed to spatiotemporal embedding. The coordinates for the $0.04^\circ$ inputs are scaled by a factor of four, whereas the $0.01^\circ$ inputs use the native grid coordinates. Details are provided in {Supplementary Note~1.4}. This encoding provides the network with the target spatial scale. The subsequent fine-tuning step then recalibrates the decoded physical fields \((\boldsymbol{v},\boldsymbol{\kappa},s)\), allowing them to adapt to the \(0.01^\circ\) grid. The numerical, training and inference configurations are summarized in {Supplementary Table 1}.}

\subsection*{Evaluation settings}

We evaluate the predictive skill and potential operational utility of Nowcast3D against three types of competitive baselines including extrapolation-based, purely data-driven, and hybrid nowcasting models (pySTEPS \cite{pulkkinen2019pysteps}, SimVP \cite{gao2022simvp}, and NowcastNet \cite{zhang2023skilful}, respectively). Specifically, the operational convective extrapolation approach pySTEPS \cite{pulkkinen2019pysteps} {serves as a radar-extrapolation reference} widely used in national meteorological services. SimVP \cite{gao2022simvp}, a deep learning model originally developed for generic video prediction, is considered as a competitive purely data-driven baseline. NowcastNet \cite{zhang2023skilful} is included as a hybrid baseline, which integrates physics-informed priors with generative modeling and has demonstrated expert-validated operational value. 

Our evaluation framework employs three complementary metrics to jointly assess the forecast quality. First, to quantify location-aware event detection, we use the neighborhood Critical Success Index (CSI) \cite{jolliffe2011forecast}, which evaluates hits, misses and false alarms within prescribed radii at multiple reflectivity thresholds. Second, to evaluate the realism of predicted precipitation structures across spatial scales, we analyze the power spectral density (PSD) \cite{sinclair2005empirical}, comparing spectral slopes and power ratios of forecasts against observations from convective to mesoscale ranges. Third, to measure perceptual similarity and structural fidelity, we employ the Learned Perceptual Image Patch Similarity (LPIPS) metric \cite{zhang2018unreasonable}. For a fair comparison with the 2D baselines, all these three metrics are evaluated on column-maximum reflectivity fields, obtained by applying the same vertical projection to the three-dimensional Nowcast3D forecasts.

{The expert-rating study was approved by the Academic Ethics Committee of Renmin University of China, Beijing, China.}

\section*{Data availability} 
The three-dimensional radar reflectivity data used in this study were obtained from {the national three-dimensional radar reflectivity mosaic product available through} the Tianqing system of the China Meteorological Administration (CMA). Restrictions apply to the availability of these data, which were used under license for the current study and are therefore not publicly available. {CMA-affiliated units can access this product through Tianqing. Other applicants should first obtain authorization from the CMA and then contact the corresponding authors, providing the authorization, intended use, and required spatial and temporal coverage. Access and use must comply with the terms of the CMA authorization.} A demonstration dataset used in this paper is publicly accessible at \url{https://github.com/Huaguan-Chen/Nowcast3D}~\cite{chen2026nowcast3d}.
\section*{Code availability} 
We use PyTorch (\url{https://pytorch.org}) for model training. Baseline experiments use the official or publicly available implementations of NowcastNet (\url{https://codeocean.com/capsule/3935105}), pySTEPS (\url{https://github.com/pySTEPS/pysteps}) and SimVP (\url{https://github.com/A4Bio/SimVP}). The code and pre-trained weights for Nowcast3D are available at \url{https://github.com/Huaguan-Chen/Nowcast3D}~\cite{chen2026nowcast3d}.

\section*{Acknowledgment}
We sincerely thank Professor Xiaoding Yu, for his important contribution to the meteorologist evaluation process. We also extend our heartfelt thanks to the 160 anonymous meteorologists who participated in the meteorologist evaluation. This work has been deployed, trained, and tested at the Tianjin Meteorological Bureau, where the model training and dataset construction were completed on the Tianjin Meteorological Observatory.
\section*{Funding Statement}
The work is supported by the Fundamental and Interdisciplinary Disciplines Breakthrough Plan of the Ministry of Education of China (JYB2025XDXM702), the National Key R\&D Program of China (2025YFE0217100), the National Natural Science Foundation of China (62276269, 62576331, 92270118), the Beijing Natural Science Foundation (F261002), and the Strategic Priority Research Program of the Chinese Academy of Sciences (XDB0620103).
\section*{Author contributions} 
H.S. (Hao Sun) and W.H. organized and led the project. H.S. (Hao Sun), W.H., and H.C. explored and designed the model and methodology. H.C. developed the Nowcast3D framework, including model architecture, training, and fine-tuning. H.S. (Haofei Sun), Y.Y., and H.C. collected and processed the radar data, while X.S. provided the wind profiler radar data. J.T. and H.C. conducted case studies and analyzed the results. N.L. contributed to the preliminary development and conceptualization of the study. H.S. (Hao Sun) and W.H. supervised all aspects of the project. All authors participated in the writing and revision of the manuscript.
\section*{Corresponding authors} 
Hao Sun (\url{haosun@ruc.edu.cn}) and Wei Han (\url{hanwei@cma.gov.cn}).

\section*{Competing interests}
The authors declare no competing interests.

\clearpage
\setcounter{figure}{0}
\renewcommand{\theHfigure}{extended.\arabic{figure}}
\renewcommand{\figurename}{Extended Data Figure}

\begin{figure}[!htp]
\centering
\includegraphics[width=0.99\linewidth,height=0.80\textheight,keepaspectratio]{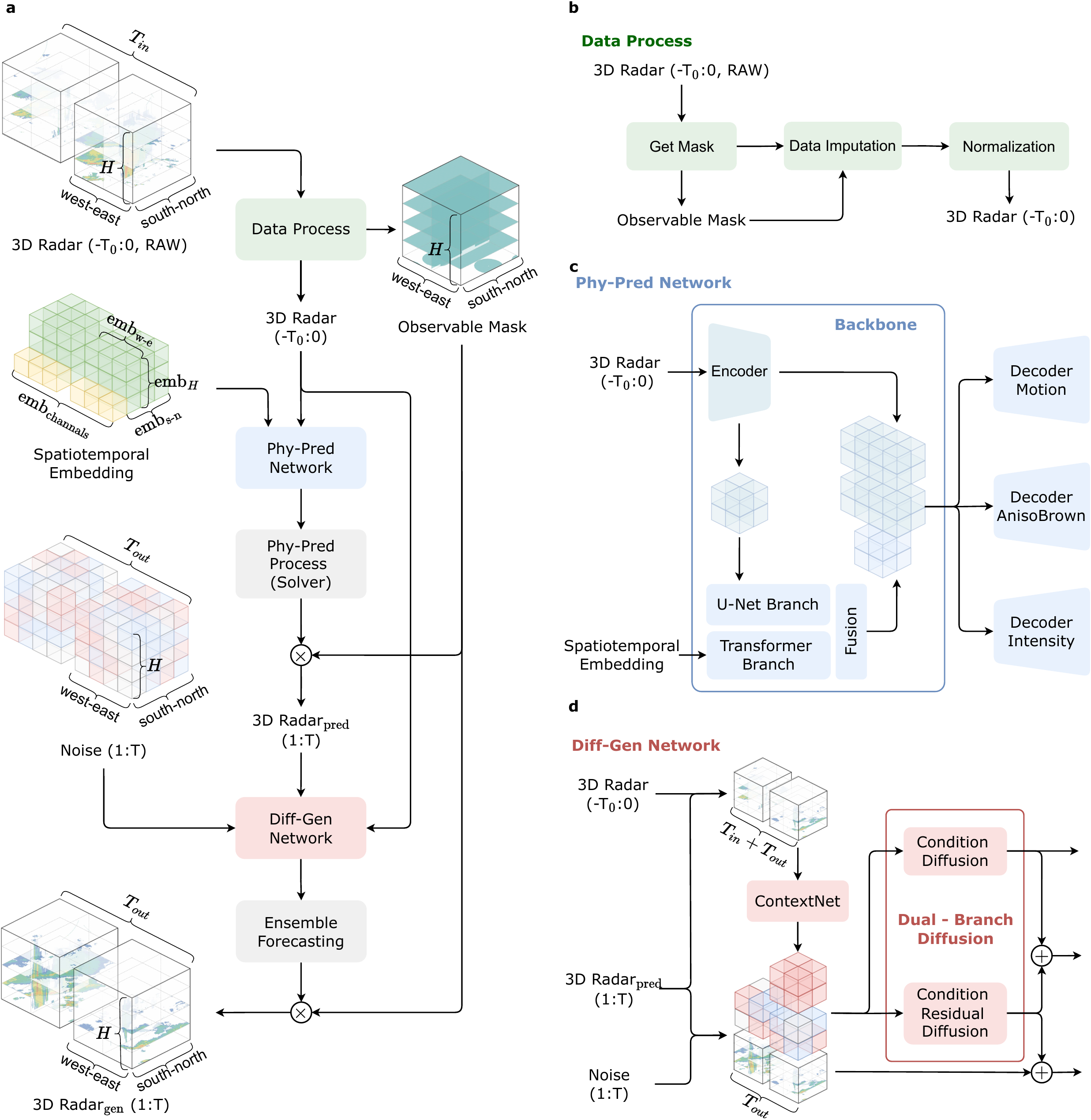}
\caption{\textbf{Overview of the Nowcast3D framework.}
The pipeline consists of three main components: data processing (green), a physics-based prediction network (blue) and a diffusion-based generative refinement network (red). \textbf{a}, Pipeline of Nowcast3D. Raw 3D radar volumes spanning the period of \textit{T}\textsubscript{in}\textit{=[-T}\textsubscript{\textit{0}}\textit{:0]} are processed into normalized inputs and an observable mask, while spatiotemporal embeddings are provided as external inputs. Phy-Pred produces a coarse forecast Radar\textsuperscript{pred} (1:\textit{T}), which is refined by Diff-Gen with injected noise to generate Radar\textsuperscript{gen}, followed by ensemble aggregation. \textbf{b}, Data processing performs mask extraction, masked imputation, and normalization. \textbf{c}, Phy-Pred fuses U-Net and Transformer branches with decoders for motion, anisotropic Brownian components, and intensity. \textbf{d}, Diff-Gen uses ContextNet and conditional diffusion on the structure and residual, conditioned on the input and Radar\textsuperscript{pred}.}
\label{fig:all_model}
\end{figure}

\clearpage

\begin{figure}[!htp]
\centering
\includegraphics[width=0.99\linewidth,height=0.80\textheight,keepaspectratio]{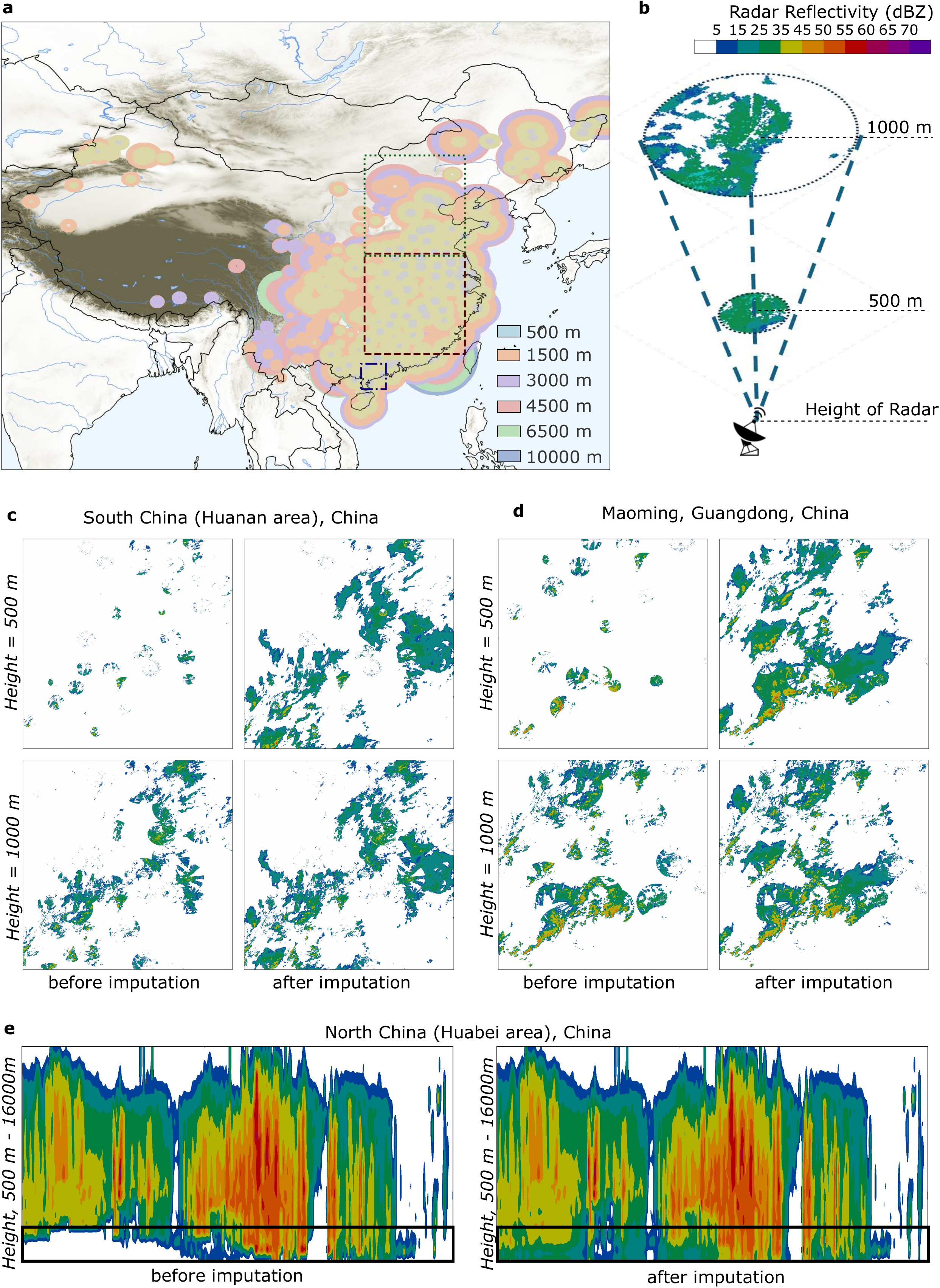}
\caption{\textbf{Radar network coverage and low-level imputation examples.}
\textbf{a}, Spatial distribution of the operational weather radar network over China and surrounding regions, with colored disks indicating approximate horizontal coverage at 500, 1500, 3000, 4500, 6500 and 10,000 m above the sea level. The dashed rectangle marks the analysis domain used in this study. \textbf{b}, Schematic of 3D radar sampling, illustrating how the beam widens with range and leads to sparse coverage near the surface despite dense sampling aloft. \textbf{c}, Example of low-level reflectivity in South China (the Huanan area) at altitudes of 500 m and 1,000 m before (left) and after (right) imputation. \textbf{d}, Same as \textbf{c} but for a convective event near Maoming, Guangdong Province, China, highlighting the recovery of coherent low-level structures beneath well-observed upper-level echoes. \textbf{e}, Height-resolved reflectivity cross-section over North China (the Huabei area) from 500 to 16,000 m before (left) and after (right) imputation, showing that the procedure completes low-level missing observations while preserving the observed vertical echo structure aloft.}
\label{fig:radarcover}
\end{figure}

\clearpage

\begin{figure}[!htp]
\centering
\includegraphics[width=0.99\linewidth,height=0.80\textheight,keepaspectratio]{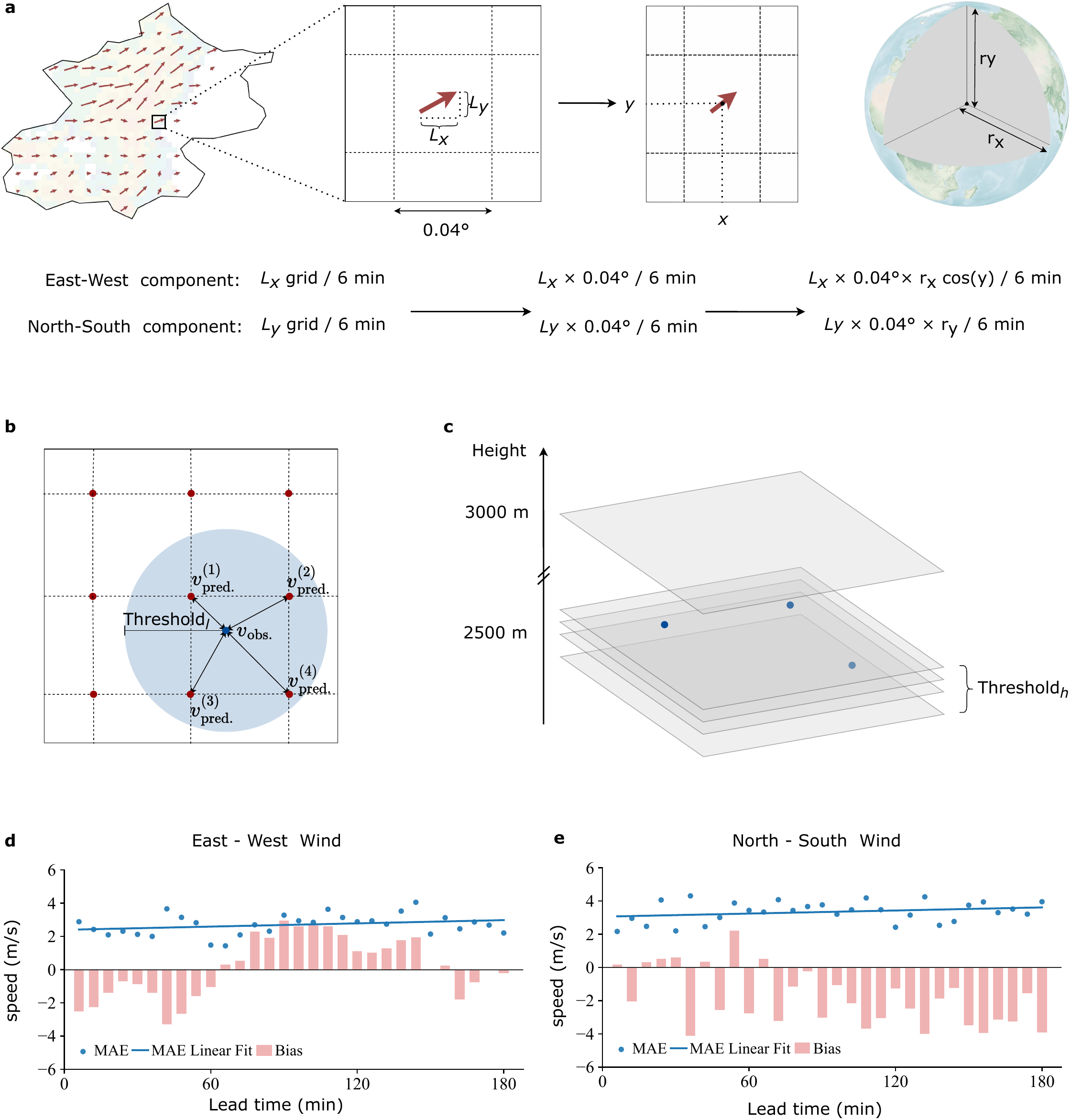}
\caption{\textbf{Wind field estimation of Nowcast3D forecast.}
\textbf{a}, Schematic of calculation from gridded data to site-specific wind field. \textbf{b}, Spatial averaging for evaluation: because observation sites do not exactly align with grid points, we define a circular area with a radius of Threshold\textsubscript{\textit{l}} around each site. The predicted values inside this circle are averaged for comparison with the site's observation. \textbf{c}, Vertical averaging for evaluation: as observations may not match the exact vertical levels of the predictions, we average inferred velocity vectors within a vertical range of \ensuremath{\pm}Threshold\textsubscript{\textit{h}}/2 at a specific height for comparison. \textbf{d}, Temporal evolution of the mean absolute error and bias in east-west wind component. \textbf{e}, Temporal evolution of the mean absolute error and bias in north-south wind component.}
\label{fig:wind_result}
\end{figure}

\clearpage

\begin{figure}[!htp]
\centering
\includegraphics[width=0.99\linewidth,height=0.80\textheight,keepaspectratio]{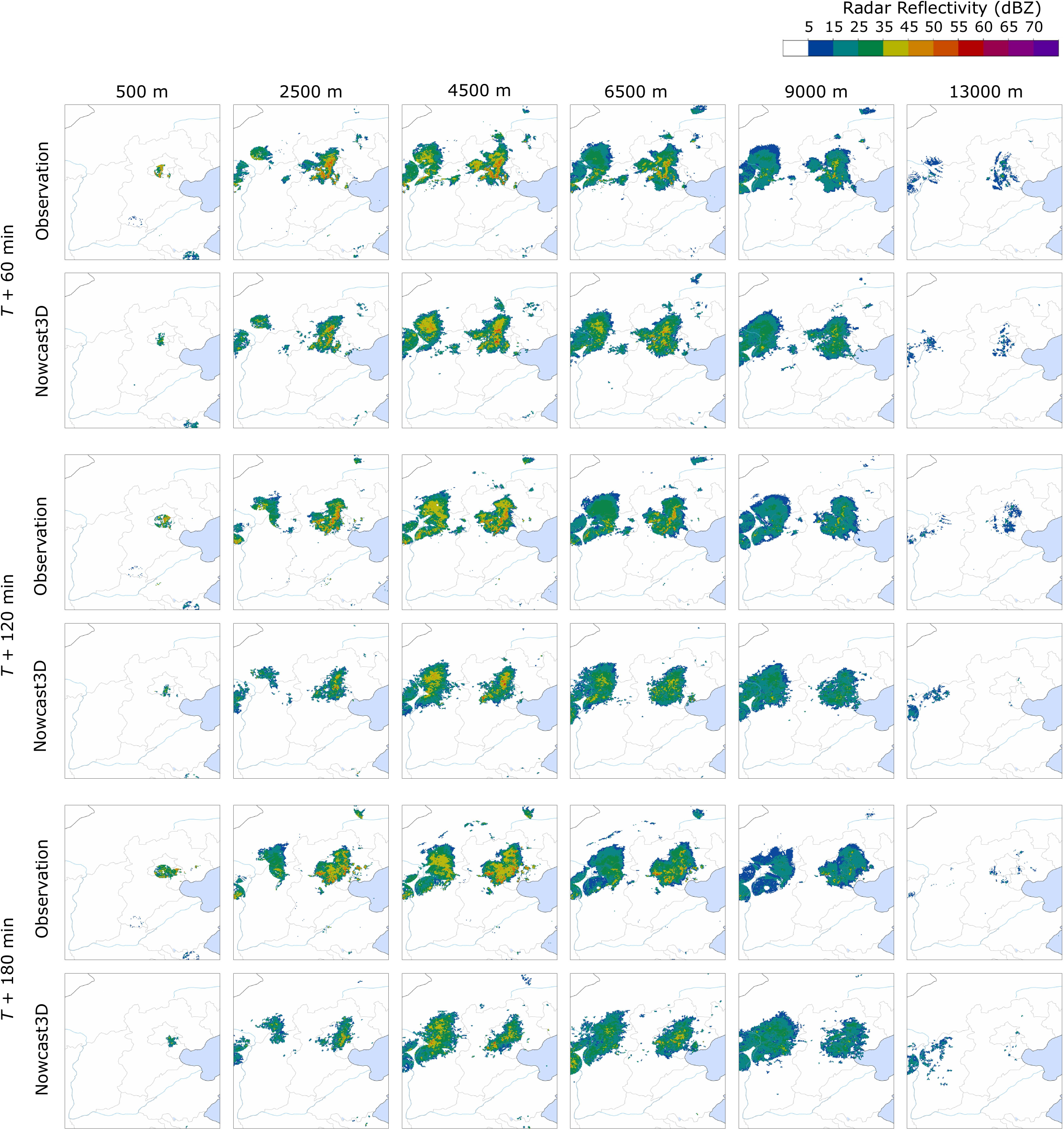}
\caption{\textbf{Nowcast3D's layered forecasts of an extreme precipitation event in North China at 60, 120, and 180 min lead times, for altitudes ranging from 500 m to 13,000 m.}
Layered prediction maps correspond to Figure 2. Columns denote altitude. Paired rows are grouped by lead time (60, 120 and 180 min), with observations followed by Nowcast3D forecasts in each pair. Reflectivity is shown in dBZ using a common color scale across panels. The observation at 500 m exhibits circular features due to the limitation of radar detection range, which appears as an overlap of multiple observation circles at low altitudes.}
\label{fig:beijing_D}
\end{figure}

\clearpage

\begin{figure}[!htp]
\centering
\includegraphics[width=0.99\linewidth,height=0.80\textheight,keepaspectratio]{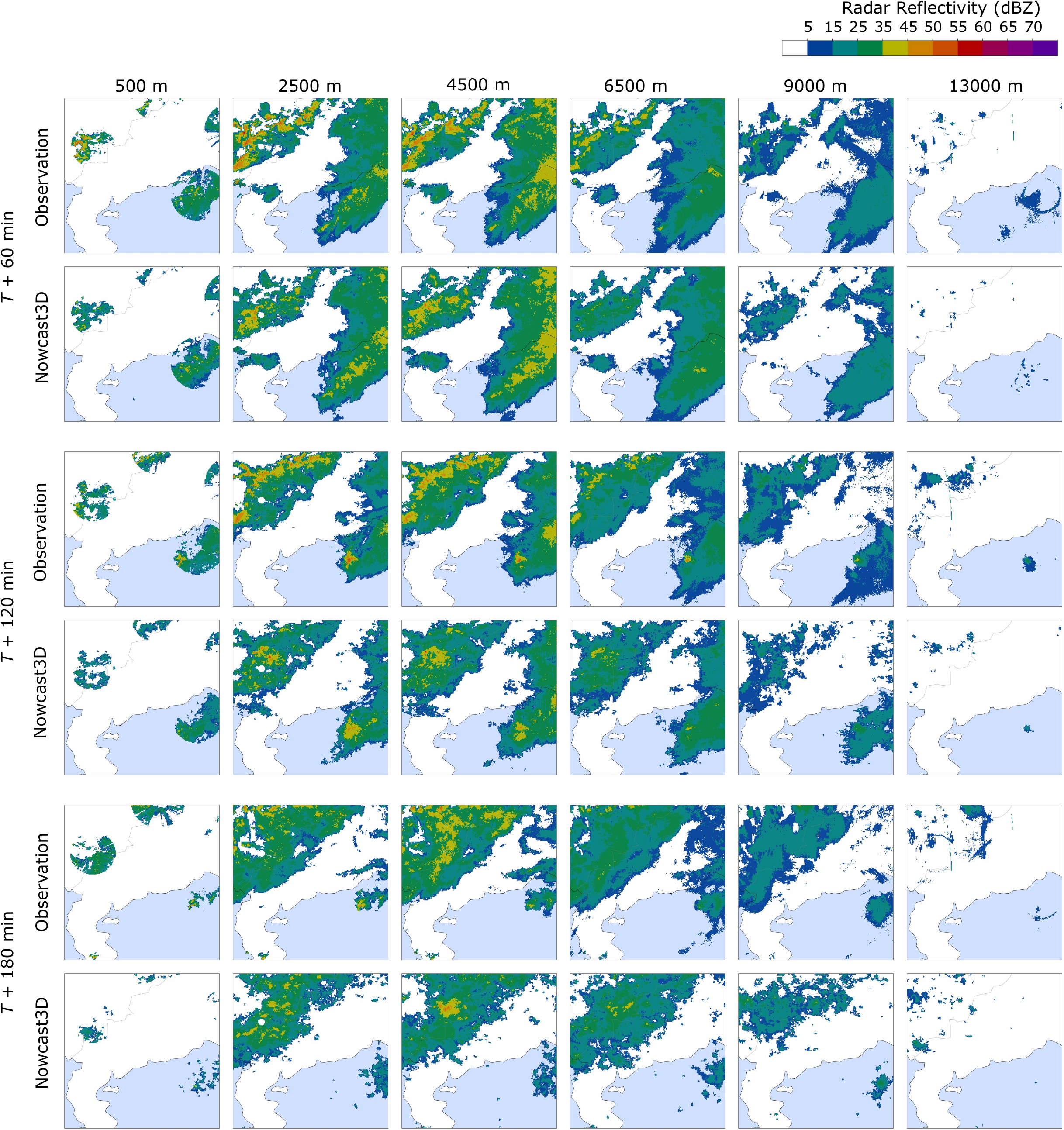}
\caption{\textbf{Nowcast3D's layered forecasts of an extreme precipitation event in Maoming, Guangdong Province, China at 60, 120, and 180 min lead times, for altitudes ranging from 500 m to 13,000 m.}
Layered prediction maps correspond to Figure 3. Columns denote altitude. Paired rows are grouped by lead time (60, 120 and 180 min), with observations followed by Nowcast3D forecasts in each pair. Reflectivity is shown in dBZ using a common color scale across panels. The observation at 500 m exhibits circular features due to the limitations of the radar detection range, which appears as an overlap of multiple observation circles at low altitudes.}
\label{fig:maoming_D}
\end{figure}

\end{document}